\documentclass[letterpaper, 10 pt, journal, twoside]{IEEEtran}
\usepackage{cite}
\usepackage{amsmath,amssymb,amsfonts}
\usepackage{multirow}
\usepackage{graphicx}
\usepackage{textcomp}
\usepackage{xcolor}
\usepackage{times}
\usepackage{soul}
\usepackage{url}
\usepackage{graphicx}
\usepackage{amsmath}
\usepackage{subfigure}
\usepackage{float}
\usepackage{bm}
\usepackage{amsmath,amssymb,amsfonts,amsthm}
\usepackage{booktabs}
\usepackage{graphicx}
\usepackage{amssymb}
\usepackage{amsmath}
\usepackage{amsthm}
\usepackage{booktabs}
\usepackage[ruled]{algorithm2e} 
\usepackage{algorithmicx}
\usepackage{algpseudocode} 
\usepackage{graphicx}
\usepackage{subfigure}
\usepackage{color,xcolor}
\ifCLASSINFOpdf
\else
\fi
\begin{document}

\title{When Transformer Meets Robotic Grasping: Exploits Context for Efficient Grasp Detection}
%


%\author{Michael~Shell,~\IEEEmembership{Member,~IEEE,}
%        John~Doe,~\IEEEmembership{Fellow,~OSA,}
%        and~Jane~Doe,~\IEEEmembership{Life~Fellow,~IEEE}% <-this % stops a space
\author{Shaochen Wang, Zhangli Zhou, and Zhen Kan,~\IEEEmembership{Senior Member,~IEEE}% <-this % stops a space
\thanks{Manuscript received February 23, 2022; revised April 25, 2022; accepted June 20,  2022.  This letter was recommended for publication by  Markus Vincze upon evaluation of the Associate Editor and Reviewers’ comments.
This work was supported in part by the National Natural Science Foundation of China under Grant U2013601, and Grant 62173314. (Corresponding author: Zhen Kan.)}
\thanks{Shaochen Wang, Zhangli Zhou,  and Zhen Kan are with the Department of Automation, University of Science and Technology of China, Hefei 230026, China, 
 (e-mail: samwang@mail.ustc.edu.cn; zzl1215@mail.ustc.edu.cn; zkan@ustc.edu.cn.)
 An extended version is available at https://arxiv.org/abs/2202.11911.
}% <-this % stops a space
\thanks{Digital Object Identifier (DOI): see top of this page.}
}

\markboth{IEEE Robotics and Automation Letters. Preprint Version. JUNE 2022}
{Wang \MakeLowercase{\textit{et al.}}: When Transformer Meets Robotic Grasping} 

\maketitle
%\thispagestyle{empty}
%\pagestyle{empty}




\begin{abstract}
	
In this paper, we present  a transformer-based architecture, namely TF-Grasp, for robotic grasp detection. The developed TF-Grasp framework has two elaborate designs making it well suitable for visual grasping tasks. The first key design is that we adopt the local window attention to capture local contextual information and detailed features of graspable objects. Then, we apply the cross window attention to model the long-term dependencies between distant pixels. Object knowledge, environmental configuration, and relationships between different visual entities are aggregated for subsequent grasp detection. The second key design is that we build a hierarchical encoder-decoder architecture with skip-connections, delivering shallow features from the encoder to decoder to enable a multi-scale feature fusion. Due to the powerful attention mechanism,  TF-Grasp can simultaneously obtain the local information (i.e., the contours of objects), and model long-term connections such as the relationships between distinct visual concepts in clutter. Extensive computational experiments demonstrate that  TF-Grasp achieves competitive results versus state-of-art grasping convolutional models and attains a higher accuracy of $97.99 \%$ and $94.6\%$ on Cornell and Jacquard grasping datasets, respectively.
Real-world experiments using a 7DoF Franka Emika Panda robot also demonstrate its capability of grasping unseen objects in a variety of scenarios.
 The code   is available at  \url{https://github.com/WangShaoSUN/grasp-transformer}.
\end{abstract}


% \fi
\IEEEpeerreviewmaketitle

\section{INTRODUCTION}

\IEEEPARstart{D}{ata}-driven methodologies such as deep learning  have become the mainstream methods for robotic visual sensing tasks such as indoor localization \cite{9672748}, trajectory prediction\cite{9309403}, and robotic manipulation\cite{kumra2017robotic},\cite{zhu2021learn}, since they require less handcrafted feature engineering and can be extended to many complex tasks. 
In recent years, as visual sensing is increasingly being used in manufacturing, industry, and medical care, growing research is devoted to developing advanced robot’s perception abilities. A typical application of visual sensing is the robotic grasp detection, where the images of objects are used to infer the grasping pose. Considering a grasping task of manipulating a wide diversity of objects, to find the graspable regions, the robots have to concentrate on not only  partial  geometric information but also the entire visual appearance of the object. Particularly in unstructured and cluttered environments, dealing with variations in shape and position (e.g., occlusion) and also the spatial relationship with other objects are critical to the performance of grasp detection.
Therefore, this work is particularly motivated to investigate grasp detection that takes into account both local neighbor pixels and long-distance relationships in spatial dimensions.

Most modern grasp detectors \cite{kumra2017robotic},\cite{redmon2015real} are based on convolutional neural networks (CNNs) which  emerge as the de facto standard for processing visual robotic grasping.  However, current CNNs are composed of individual convolution kernels, which are more inclined to concentrate on local level information.  Also, the convolution kernels in a layer of CNN are viewed as independent counterparts without mutual information fusion. 
Generally, to maintain a large receptive field, CNNs have to repeatedly stack convolutional layers, which reduce the spatial resolution and inevitably results in the loss of global details and degraded performance.

Recently, as a novel approach to handle natural language processing and computer vision, the transformer \cite{attent},\cite{DBLP:conf/iclr/DosovitskiyB0WZ21},\cite{liu2021swin} demonstrates remarkable success. The widely adopted attention mechanisms \cite{attent} of transformers in sequence modeling provide an elegant resolution that can better convey the  fusion of information across global sequences.
In fact, as robots are deployed in more and more diverse applications such as industrial assembly lines and smart home, 
the sensing capacity of  robotic systems needs to be enriched, not only in local regions, but also in global interaction.
Especially when robots frequently interact with objects in the environment, the awareness of global attention is particularly important with respect to safety and reliability. However, most vision transformers are designed for image classification on natural images processing tasks.
Few of them are specifically built for robotic tasks.

In this paper, we present a transformer-based visual grasp detection framework, namely TF-Grasp,  which leverages the fact that the attention can better aggregate information across the entire input sequences to obtain an improved global representation. More specifically, the information within  independent image patches is bridged via self-attention and the encoder in our framework  captures these multi-scale low-level features. The decoder incorporates the high-level features through long-range spatial dependencies to construct the final grasping pose.
We provide detailed empirical evidence to show that our grasping transformer  performs reasonably well on popular grasping testbeds,  e.g., Cornell and Jacquard grasping datasets.
The experimental results demonstrate that the transformer architecture plays an integral role in generating appropriate grasping poses by learning local and global features from different parts of each object. The vision transformer-based grasp detection works well on the real robotic system and shows promising generalization to unseen objects. In addition, our TF-Grasp can generate the required grasping poses for parallel grippers  in a single forward pass of the network.

In a nutshell, the contributions of this paper can be summarised in three folds:
\begin{itemize}
	\item This work presents a novel and neat transformer architecture for visual robotic grasping tasks.  To the best of our knowledge, it is one of the first attempts considering vision transformers in grasp detection tasks.
	\item We consider simultaneous fusion of local and global features and redesign the classical ViT framework for robotic visual sensing tasks.
	
	\item Exhaustive experiments are conducted to show the advantages of the transformer-based robotic perception framework.
	The experimental results demonstrate that our model achieves improved performance on popular grasping datasets compared to the state-of-the-art methods. We further show that our grasping transformer can generate appropriate grasping poses for known or unknown objects in either single or cluttered environments.
\end{itemize}

\section{Related work}
This section reviews recent advances in the field of robotic grasping and briefly describes the progress of transformers in different areas.

\subsection{Grasp Detection}
The ability to locate the object position and determine the appropriate grasping pose is crucial to stable and robust robotic grasping. Grasp detection, as the name implies, uses the image captured from the camera to infer the grasping pose for the robot manipulator. Using geometry-driven methods, earlier works \cite{murray2017mathematical}, \cite{DBLP:conf/icra/BicchiK00} mainly focus on analyzing the contours of objects to identify grasping points. A common assumption in these methods is that the geometric model of the object is always available. However, preparing the CAD models for graspable objects is time-consuming and impractical for real-time implementation. Recently, deep learning based methods
have been successfully applied in visual grasping tasks \cite{kumra2017robotic}, \cite{redmon2015real}, \cite{zhang2019roi},\cite{asif2018graspnet},\cite{DBLP:conf/icra/ZhuSFT21}. The work of \cite{lenz2015deep} is one of the earliest works that introduces deep neural networks to grasp detection via a two-stage strategy where the first stage finds exhaustive possible grasping candidates and the second stage evaluates the quality of these grasp candidates to identify the best one. However, due to numerous grasping proposals, the method in \cite{lenz2015deep} suffers from relatively slow speed. Many recent works utilize convolutional neural networks to generate bounding box proposals to estimate the grasp pose of objects. 
Redmon et al.\cite{redmon2015real} employed an  Alexnet-like CNN architecture to regress grasping poses. Kumra et al.\cite{kumra2017robotic} explored the use of ResNet-50 as a backbone to  incorporate multimodal including depth and RGB information to further improve the  grasp performance. Besides, CNN-based grasp quality networks \cite{DBLP:conf/rss/MahlerLNLDLOG17},\cite{gariepy2019gq}  were proposed to evaluate and predict the robustness of grasp candidates.  In the same line, GG-CNN \cite{morrison2020learning} developed  a fully convolutional neural network to perform grasp detection, which provides a lightweight and real-time solution for visual grasping.
Currently, most of the existing grasp detection methods are still heavily inspired by computer vision  techniques such as object recognition, object detection, etc. In contrast to classical visual problems where the detected objects are usually well-defined instances in the scene, in grasp  detection, the grasp  configuration to be generated is continuous, which implies an infinite number of possible grasp options.
This places significant challenges in feature extraction to identify a valid grasp configuration from all possible candidates. We argue that the loss of long-term dependencies in feature extraction is a major drawback of current CNNs based grasp detection methods.

\subsection{Transformer}
Transformer \cite{attent} first emerged in machine translation and  is rapidly establishing itself as a new paradigm in natural language processing due to  its potential to model global information, which learns the high  quality features by considering the whole context.  Thanks to its excellent global representation and friendly parallel computation, the transformer is competitive in long sequences modeling and gradually replaces RNNs and CNNs. 

Motivated by the remarkable success of transformers achieved in natural language processing, more and more researchers are interested in the employment of attention mechanisms in visual tasks. At present,  the transformer has been successfully applied to image classification, object detection, and segmentation tasks.   
However, there  still exist many challenges.
First, visual signals and word tokens are very different on many scales.  Second, the high dimension of pixel-level information may introduce significant computational complexity.

More recently, ViT \cite{DBLP:conf/iclr/DosovitskiyB0WZ21} was presented as a transformer model to tackle natural images recognition, which splits the image into non-overlapping patches. The authors in \cite{liu2021swin} proposed a hierarchical ViT called Swin-Transformer by calculating the local self-attention with shifted windows.
In contrast to the quadratic computation complexity of self-attention in ViT, Swin-Transformer achieves a linear complexity. Inspired by this fashion, many researchers have tried to apply transformer to other fields. For example, TransUNet \cite{chen2021transunet} combines transformer and Unet \cite{ronneberger2015u} for medical image diagnosis.
Nevertheless, how to exploit the strengths of attention to aggregate information from entire inputs  has not been investigated in the task of visual grasp detection.
Unlike prior works, we design a transformer based encoder-decoder architecture to predict the grasp posture in an end-to-end manner. It is shown that our method achieves higher grasp success than the state-of-the-art CNNs counterparts.

\section{Method}

\textbf{Grasp Representation}. 
The autonomous visual grasping tasks generally start from collecting visual images of the object by sensory input, which will then be processed to generate an effective grasp configuration to maximise the probability of grasp success. Considering a parallel-plate gripper, the grasp representation $g$ \cite{jiang2011efficient} is formulated as a 5-dimensional tuple:
\begin{equation} 
    g=\{x,y,\theta,w,h\}
\end{equation}
where $(x,y)$ are the center coordinates of the grasp rectangle, $(w,h)$ denote the  width and height of the grasp rectangle, and $\theta$ is the orientation of the grasp rectangle with respect to the horizontal axis. Given a gripper with known dimensions, a simplified representation can be expressed as $g=(p, \phi, w)$ where $p=(x,y)$, $\phi$ indicates the orientation angle of gripper and $w$ denotes the opening distance of gripper, respectively.

To facilitate grasping, we follow the setting in \cite{morrison2020learning} to represent the grasp in 2-D image space as 
\begin{equation}
	G=\{Q, {W},{\Theta} \} \in \mathbb{R}^{3 \times W \times H},
\end{equation}
where the grasp quality $Q$ measures the grasp success of each pixel, and $W$ and $\Theta$ are the gripper width and orientation angle maps. The value of each pixel in $W$ and $\Theta$ represents the corresponding width and angle of gripper at that position during the grasping.

\begin{figure}[]
	
	\center
	\subfigure{
		\includegraphics[width=0.32\textheight]{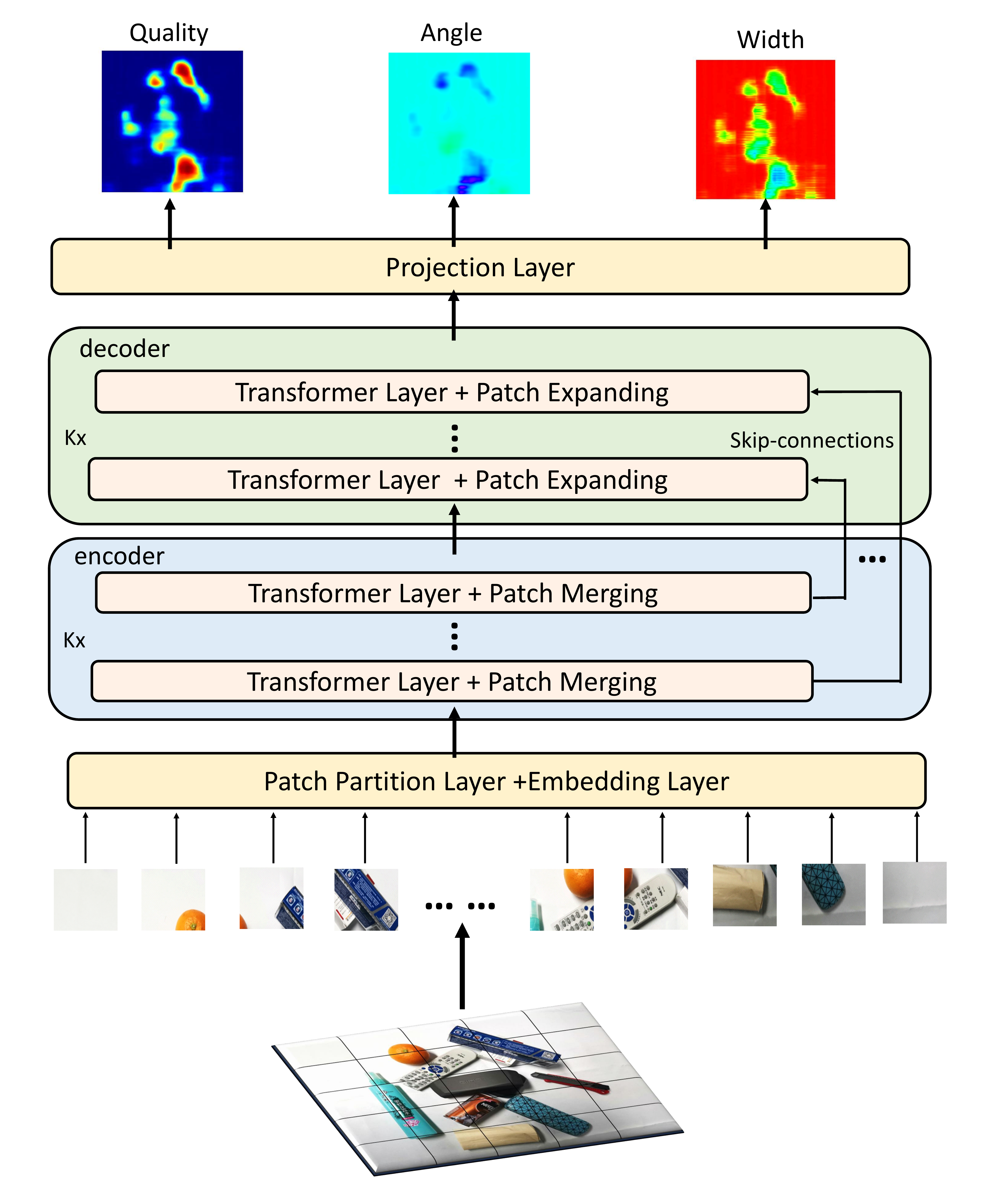}
	}
	\caption{ Overview of the TF-grasp model. Our model takes as input the image captured by the camera mounted on the end-effector of the manipulator and generates a pixel-level grasp representation. }	
	\label{arch}
\end{figure}

Consequently, in the developed TF-Grasp, the grasp detection task boils down to three sub-tasks, namely the problems of predicting grasping position, angle, and width.

{\textbf{Grasp Transformer Overview}.}
A deep motivation of this work is that  the treatment of robot  perception in complex, dynamic robotic tasks should be global and  holistic with information mutual fusion.
Specifically, the grasping model can be formulated into an encoder-decoder architecture with a U-shaped structure, as detailed in Fig. \ref{arch}. The encoder branch  aggregates the entire visual input, mutually fuses features by using attention blocks, and then  extracts the specific features that are useful for visual robotic grasping.  During the decoder process,  the model  incorporates features delivered via skip-connections and performs a pixel-level grasp prediction by up-sampling. 
More concretely, the attention modules in the decoder enable more comprehensive processing of local and long-range information, allowing  for better multi-scale feature fusion. 
Each pixel in the  prediction heatmap is correlated with the final location and orientation of the end-effector. 

To bridge the domain gaps between the transformer and visual robotic grasping tasks, {we have carefully designed our grasping transformer in the following aspects for improved grasp detection. (a) Cascade Design. Different from the classic ViT architecture, we adapt a cascaded encoder-decoder structure. The encoder utilizes self-attention to learn a contextual representation that facilitates grasping and the decoder makes use of the extracted features to perform a pixel-level grasp prediction.  (b) Local and Global balance. We utilize the swin attention layer to achieve a trade-off between global and local information for better scene perception. Window attention performs local feature extraction  and the shifted-window attention  allows cross window interactions to globally focus on more diverse regions.      (c) Feature Fusion. The feature representations at
different stages are connected by skip-connections for a multi-scale feature fusion, which acquire both rich semantic and detailed features.    (d) Lightweight  Design. It is essential for robots to account for efficiency and the real-time performance. We utilize shifted attention blocks and a slimming design  for our grasping transformer to reach an ideal trade-off between the performance and speed.
}

\textbf{Grasp Transformer Encoder}. 
Before being fed  into the encoder, the image is first passed through patch partition layer and is then cut into non-overlapping patches.  Each patch is treated as a word token in the text. For example,
a 2D image $I \in \mathbb{R}^{W \times H \times C}$ is split into fixed-size patches $x \in \mathbb{R}^{N \times (P \times P \times C )}$, where $(H, W)$ denote the height and width of the original image, $C$ represents the channel of the image, $P$ is the shape size of each image patch, and $N=H \times W / P^2$ refers to the number of image patches. Then token-based representations can be obtained by passing the images patches into a projection layer. 

The encoder is composed by stacking identical transformer blocks. Attentions in the transformer block build long-distance interactions across  distant pixels and attend on these positions in the embedding space.  At the top of the encoder is a bottleneck block attached to the decoder.
The fundamental element in our grasping transformer framework is the multi-head self-attention.
The input feature $\mathbf{X}$  is linearly transformed to derive the query $Q$, key $K$, and value $V$, which are defined as follows:
\begin{equation}
	Q =X W_Q, K=X W_K, V=XW_V,
\end{equation}
where $W_Q,W_K,W_V$ are linear projection matrices. Next, we compute the similarity between the query and key by using the dot product to obtain the attention, 
\begin{equation}
     \text{Attention}({Q},{K},{V})= \operatorname{SoftMax}(\frac{{Q} {K}^T}{\sqrt{d}}+{B}) {V}
\end{equation}
where $\sqrt{d}$ is the scaling factor and $B$ is the learnable relative position encoding.

\begin{figure}[]
	
	\center
	\subfigure{
		\includegraphics[width=0.27\textheight]{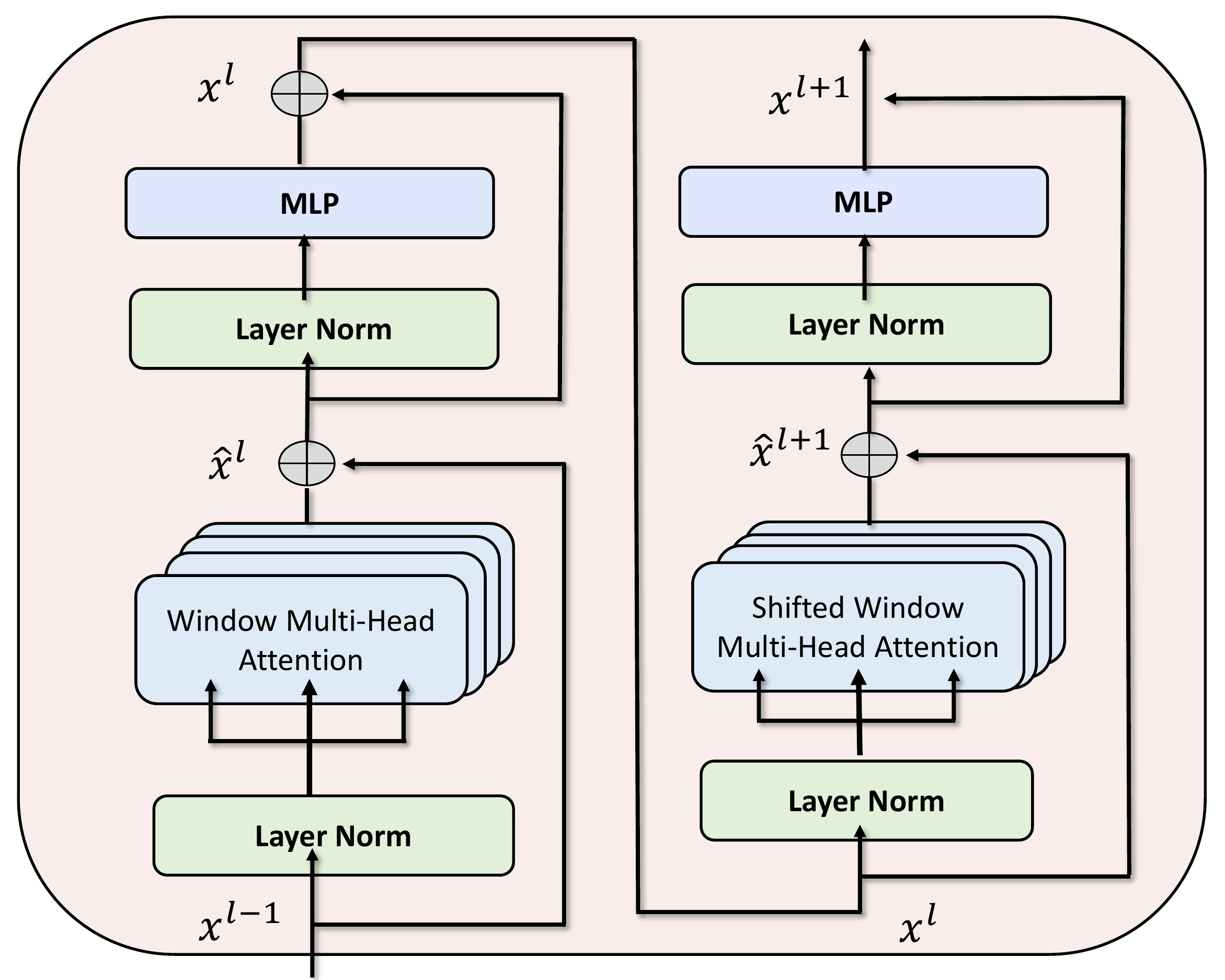}
	}
	\caption{ The architecture of our transformer block.  }	
	\label{swin}
\end{figure}

The computational complexity  of self-attention grows quadratically with respect to the image size. To achieve computational efficiency, we leverage the advantages of CNNs and transformer and adopt the swin-transformer block \cite{liu2021swin} in our framework.  The  swin-transformer layer consists of two parts:  local attention and global attention. Within the local attention, the calculation of self-attention is restricted to  local regions where images patches are divided into non-overlapping local  windows. Cross-window attention introduces connections  between neighbors by sliding non-overlapping windows. 
The structure of swin-transformer block is presented in Fig. \ref{swin} which is composed of MLP, Layer Norm, window-based MSA and shifted-window MSA.
The computation procedure of swin-transformer block is represented as follows:
\begin{equation}
\begin{aligned}
  \hat{\mathbf{x}}^{l} &=\operatorname{W-MSA}\left(\mathrm{LN}\left(\mathbf{x}^{l-1}\right)\right)+\mathbf{x}^{l-1}, \\
  \mathbf{x}^{l} &=\operatorname{MLP}\left(\mathrm{LN}\left(\hat{\mathbf{x}}^{l}\right)\right)+\hat{\mathbf{x}}^{l}, \\
  \hat{\mathbf{x}}^{l+1} &=\operatorname{SW-MSA}\left(\mathrm{LN}\left(\mathbf{x}^{l}\right)\right)+\mathbf{x}^{l}, \\
  \mathbf{x}^{l+1} &=\operatorname{MLP}\left(\mathrm{LN}\left(\hat{\mathbf{x}}^{l+1}\right)\right)+\hat{\mathbf{x}}^{l+1}
\end{aligned}
\end{equation}
where W-MSA and SW-MSA  refer to the local window and global shifted window multi-head self-attention, respectively.
$\mathbf{x}^{l-1}$ denotes the feature of output from the previous layer. Then, the features will be sent into the window attention, $ \operatorname{W-MSA}$. There is  a layer norm before both MLP and attention layer, and residual connections are applied to these modules.
Between every two swin transformer blocks, there exists a patch merging operation that reduces the resolution of feature maps.
The patch merging layer builds a hierarchical representation by gradually merging consecutive neighboring patches between successive transformer layers.

\textbf{Grasp Transformer Decoder}. 
The decoder generates an executable grasping configuration that allows the end-effector to move to the corresponding positions. We transform the planar grasp detection problem into a pixel-level prediction. Three grasping heads are attached in parallel to the top  of the decoder, including a grasp confidence head {$Q$}, a gripper angle head $\Theta$, and a gripper width head $W$.  
The output of each head is a heat map with the same size as the input visual image. The grasp confidence head outputs a value between 0 and 1, which indicates the probability of the successful grasping at each pixel point. Likewise, the gripper width and angle heads output the width and rotation angle of the gripper when grasping at the corresponding point in the image, respectively. We treat the grasping posture estimation as a regression problem  and use our transformer model to learn a mapping $F: I \rightarrow \tilde{G}$ by minimizing the distances between the predicted grasping heatmaps $\tilde{G} ({Q}, {W},{\Theta})$ and the ground truth, where   $I$ is the input data. The loss function is defined as follows:
\begin{equation} 
\mathcal{L}= \sum_i^N \sum_{m \in\{{Q}, {W},{\Theta} \}} \| \tilde{G}_i^m - L_i^m  \|^2
\end{equation}
where $N $ is the number of  sample size and $L_i$ is the corresponding label.

The ultimate grasp location is the position with the highest grasp confidence by retrieving the grasp quality heatmap, defined as:   
\begin{equation} 
\mathcal{G^{*}}_{pos}= \operatorname{argmax}_{pos} \text{Q},
\end{equation}
where $\text{Q}$ is the grasp confidence map.  Afterward, we extract the predicted angle $\theta$ and angle $w$ of the corresponding position from the angle and width heatmaps.

In our grasp detection decoder, we also adopt swin transformer block to reduce the computational complexity.  
Swin attention aggregates multi-scale features and builds a hierarchical representation. And skip-connections merge the features learned at these different stages for further fusion to produce a better grasp posture. Analogous to U-net \cite{ronneberger2015u}, skip-connections  are implemented by concatenating features from the $i$-th layer of  the encoder directly into the layer $i$-th in the decoder. 
In the decoding phase, following the patch expanding layer, the concatenated features are
taken as input to the next attention block stage. Simultaneously, we can learn the relationship between the fused features where the features in the encoder can be used as queries and keys to interact with the counterparts in the decoder for self-attention computing.

A benefit of our pixel-level grasp representation is that only a single forward propagation is required to obtain the best grasp postures within the global visual scene, avoiding the need to generate multiple grasp candidates and saving the computation expense.

\begin{figure*}[]
	\subfigure{
		\includegraphics[width=0.095\textheight]{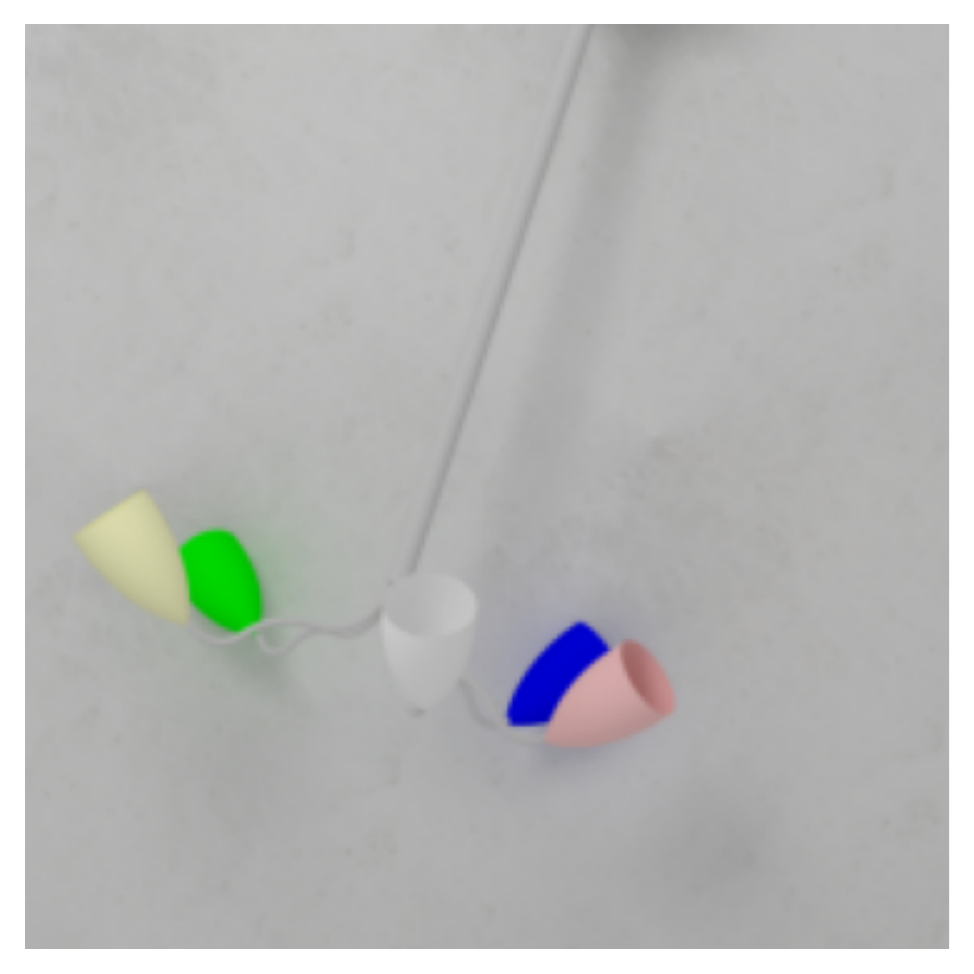}
		\includegraphics[width=0.095\textheight]{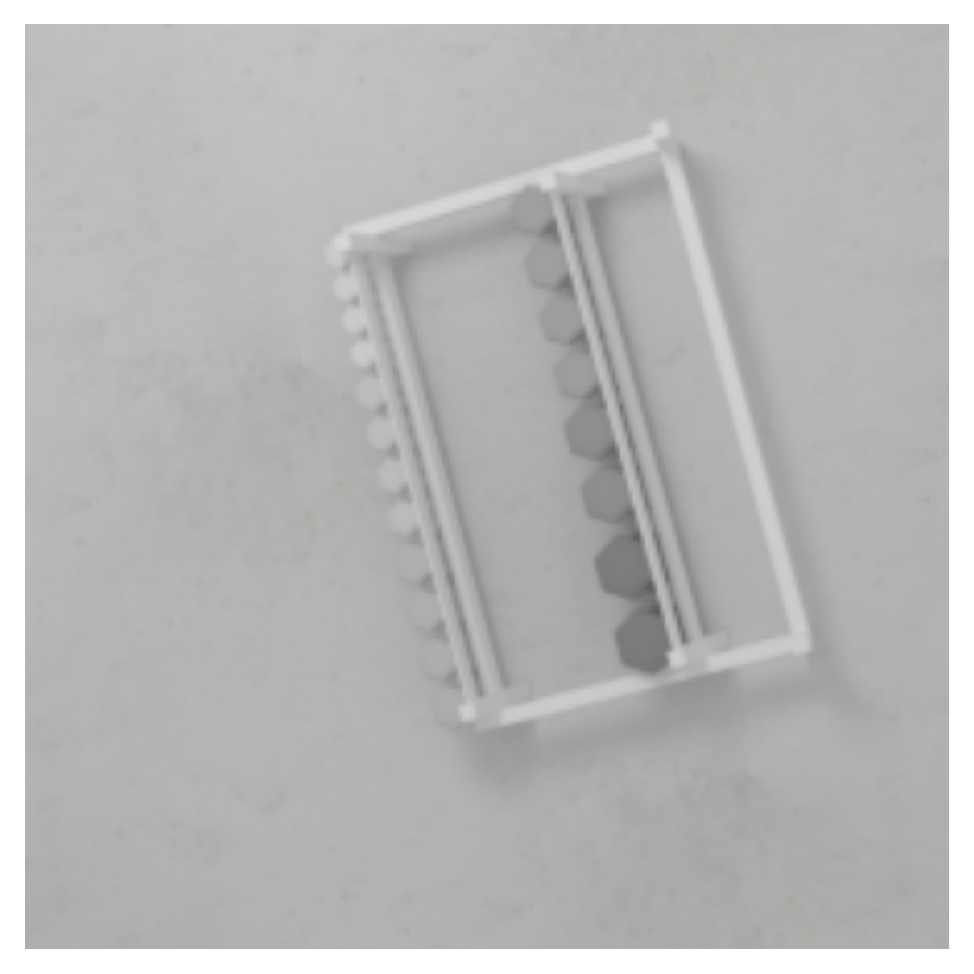}
		\includegraphics[width=0.095\textheight]{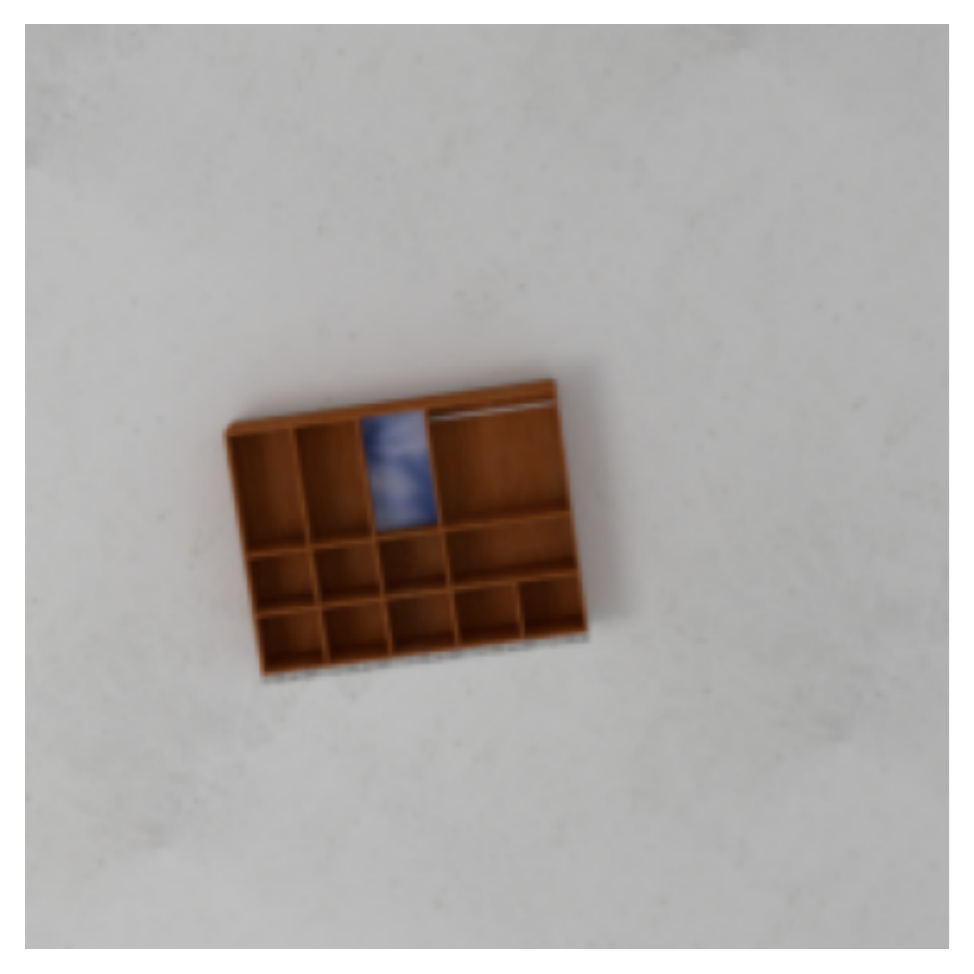}
		
		\includegraphics[width=0.095\textheight]{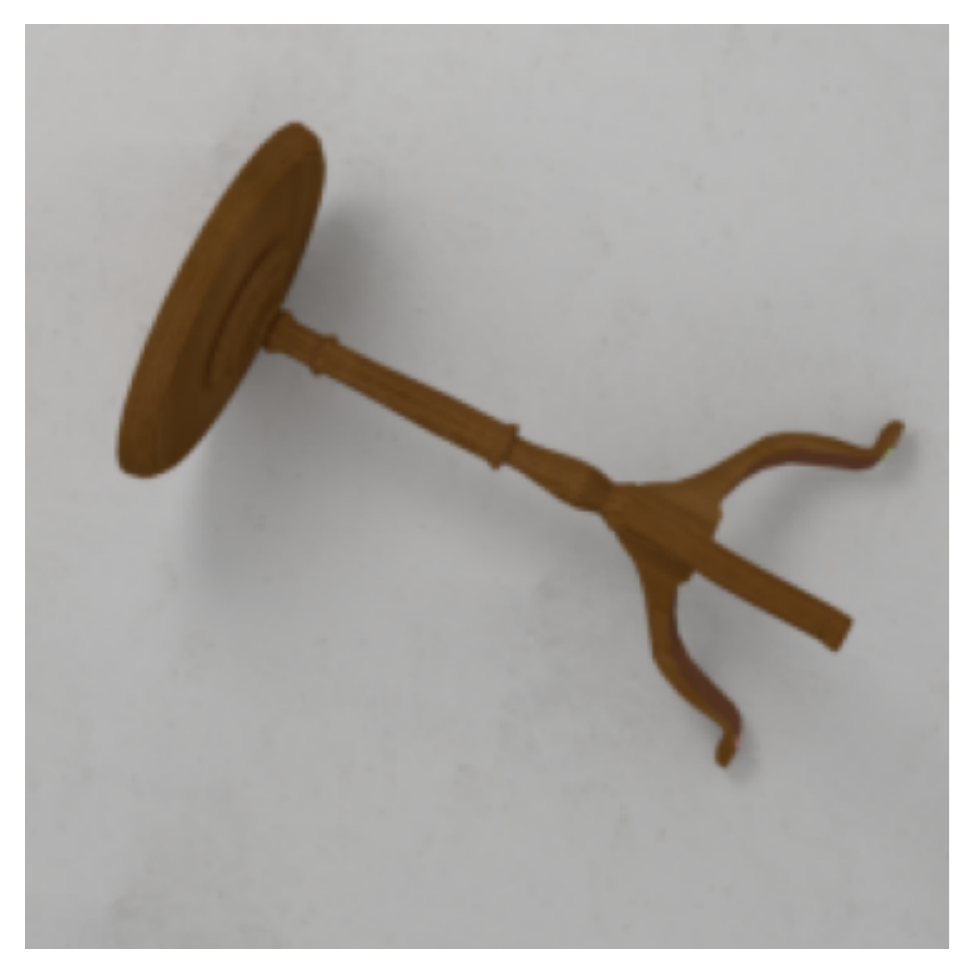}
		\includegraphics[width=0.095\textheight]{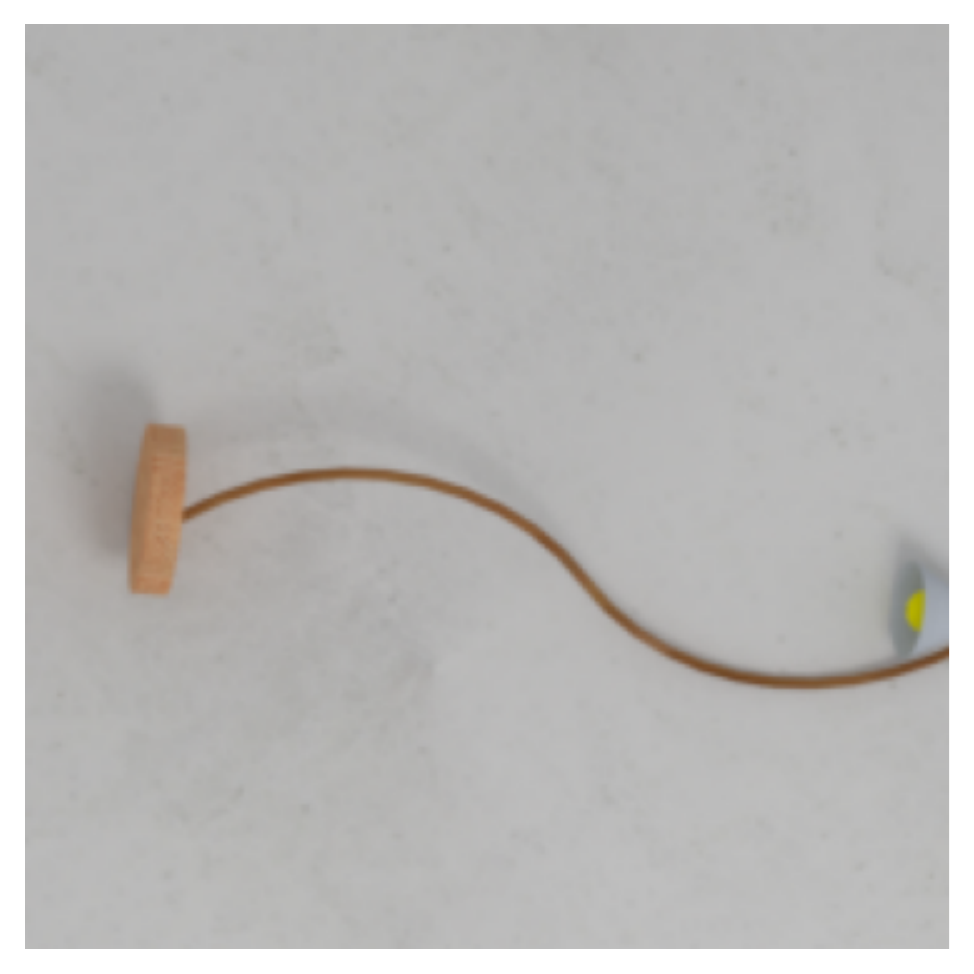}
		\includegraphics[width=0.095\textheight]{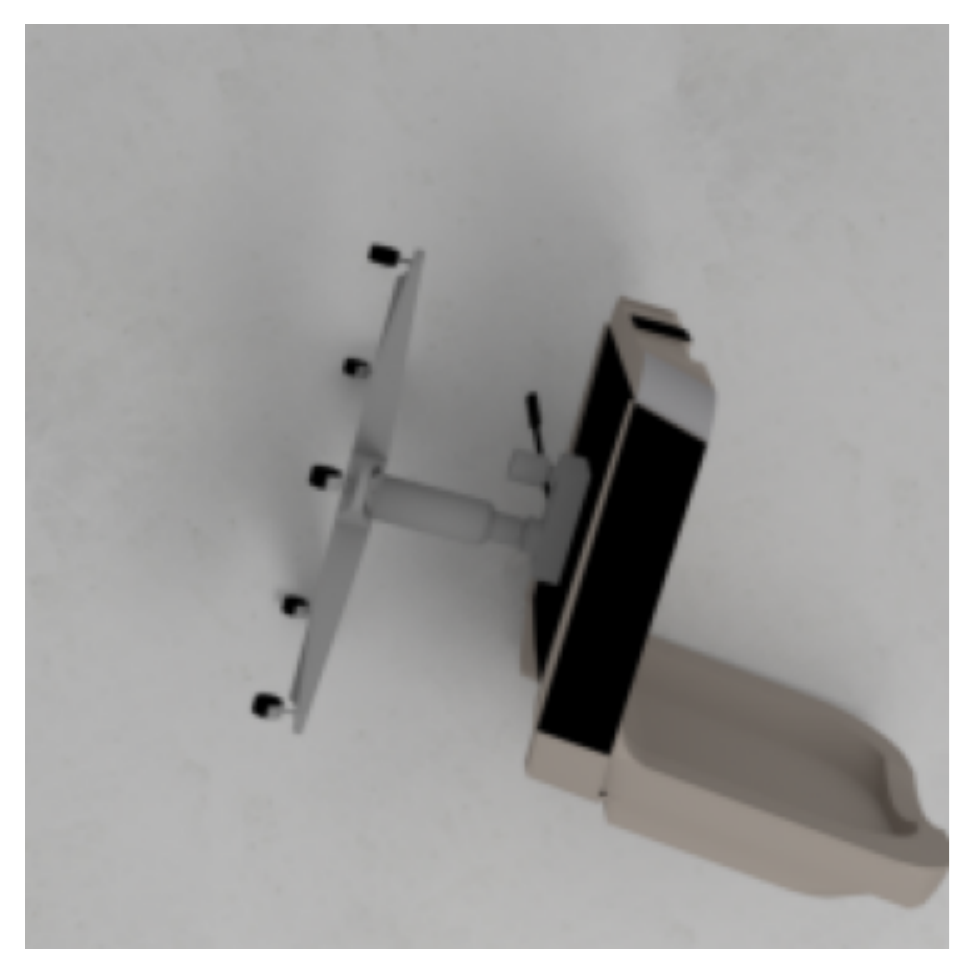}
		\includegraphics[width=0.095\textheight]{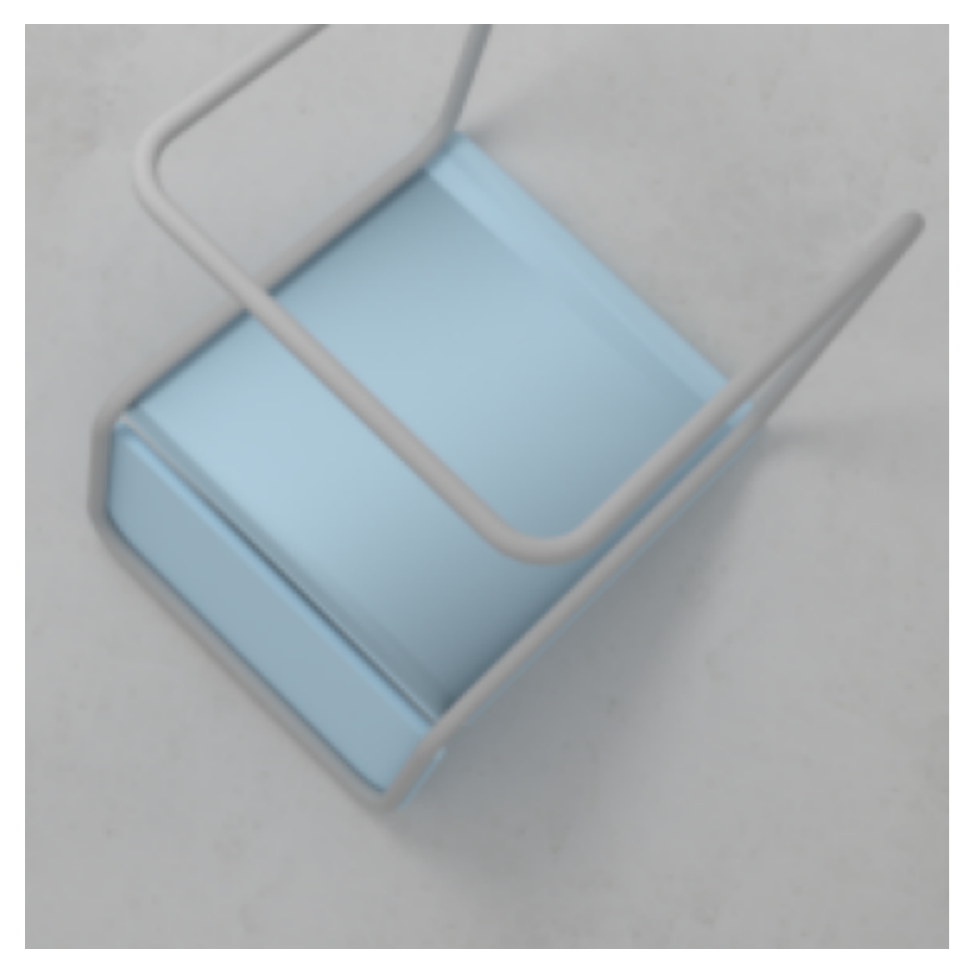}
	}
	\subfigure{
		\includegraphics[width=0.095\textheight]{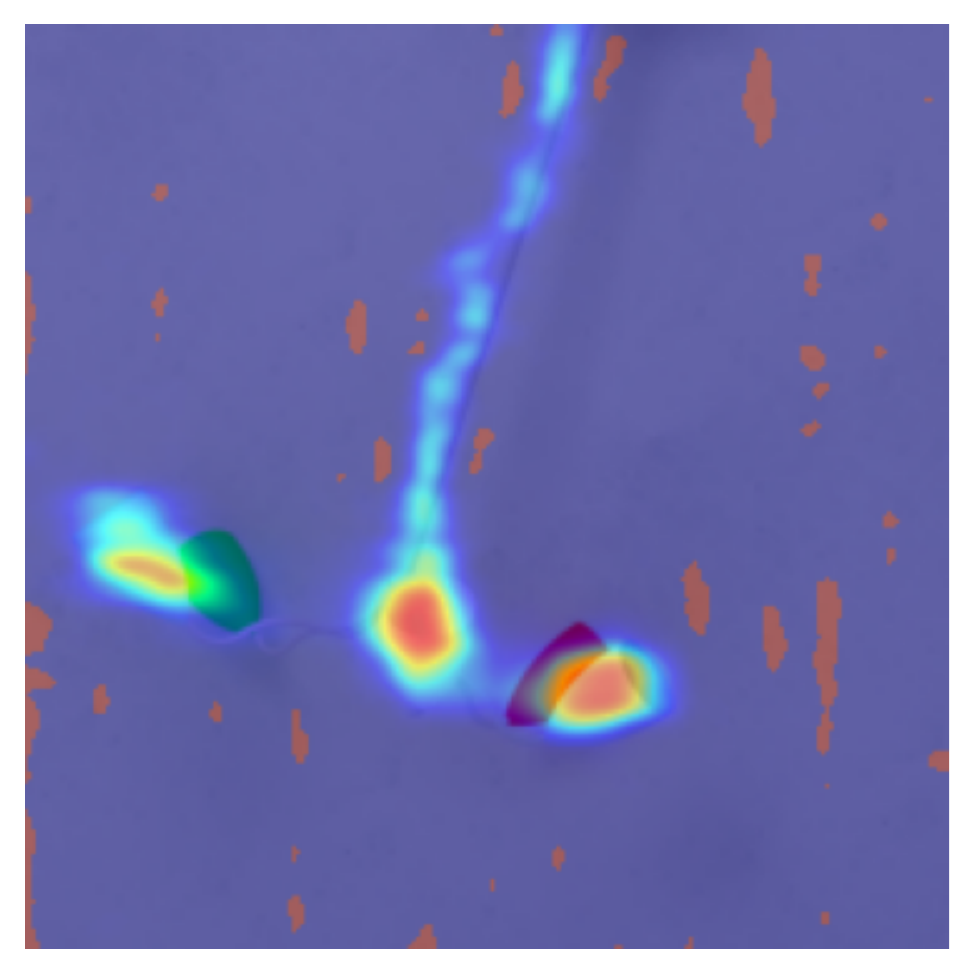}
		\includegraphics[width=0.095\textheight]{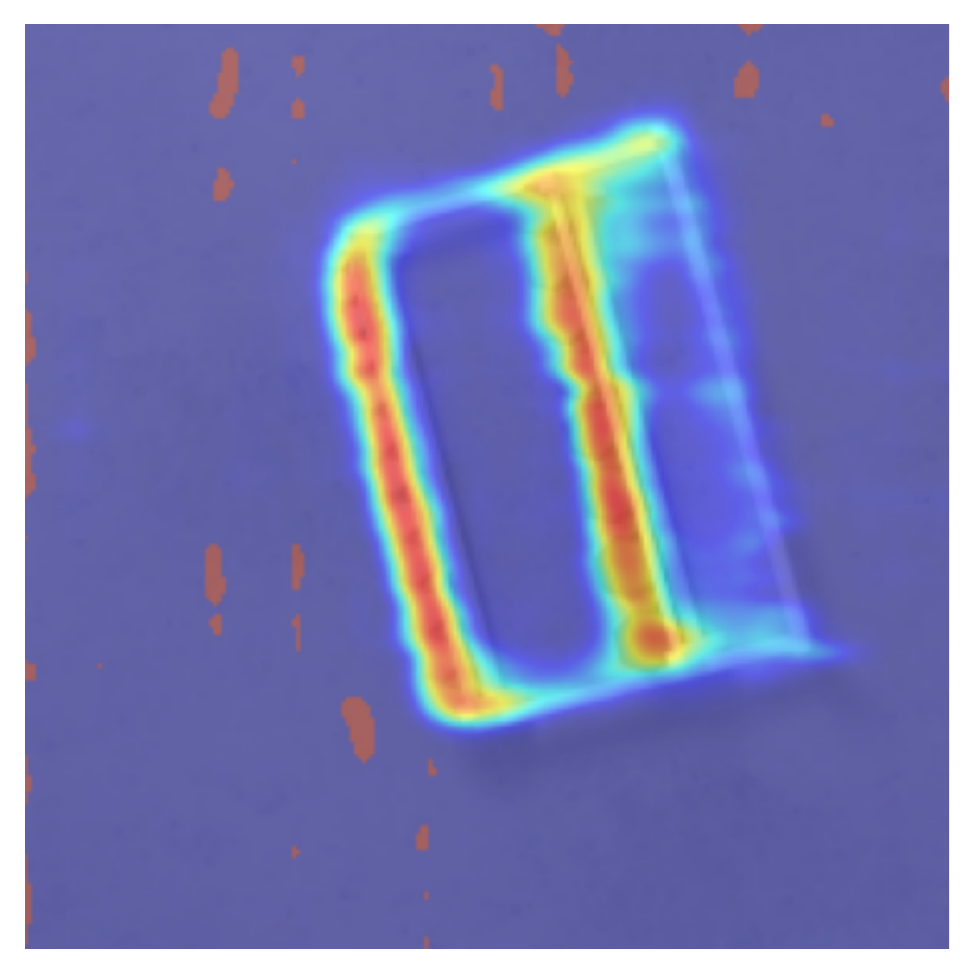}
		\includegraphics[width=0.095\textheight]{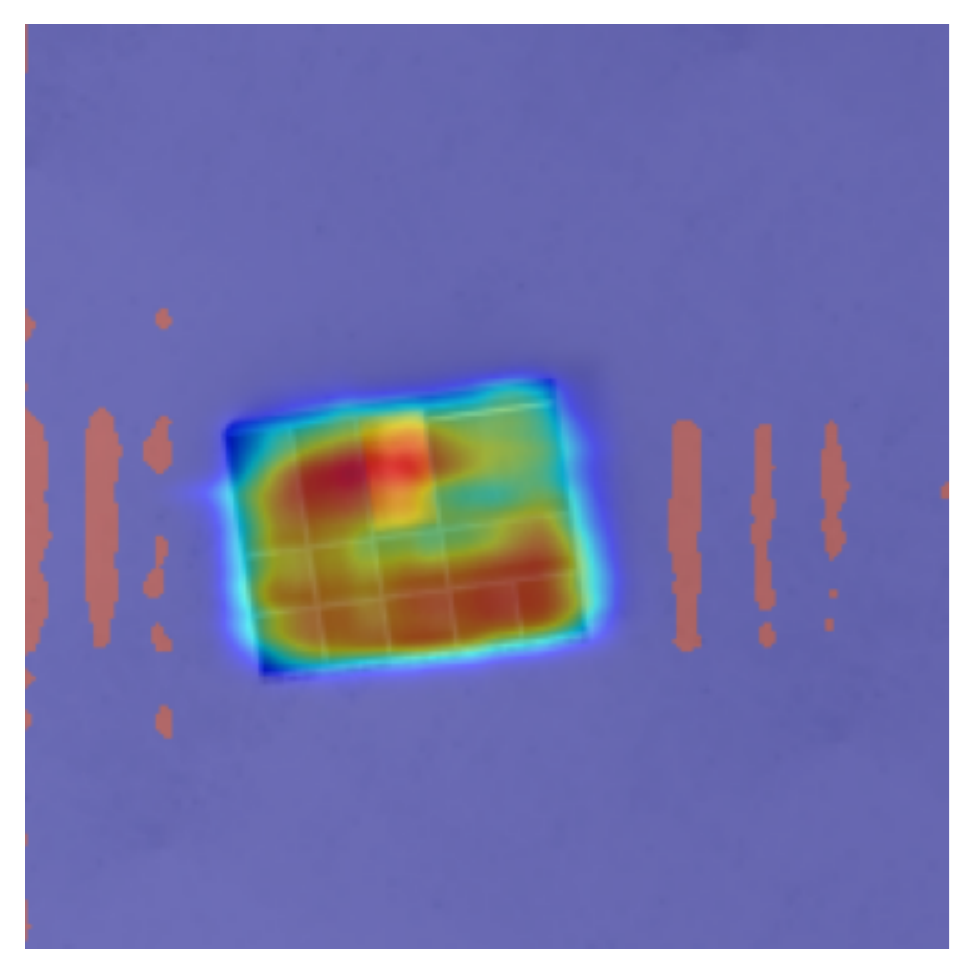}
		
		\includegraphics[width=0.095\textheight]{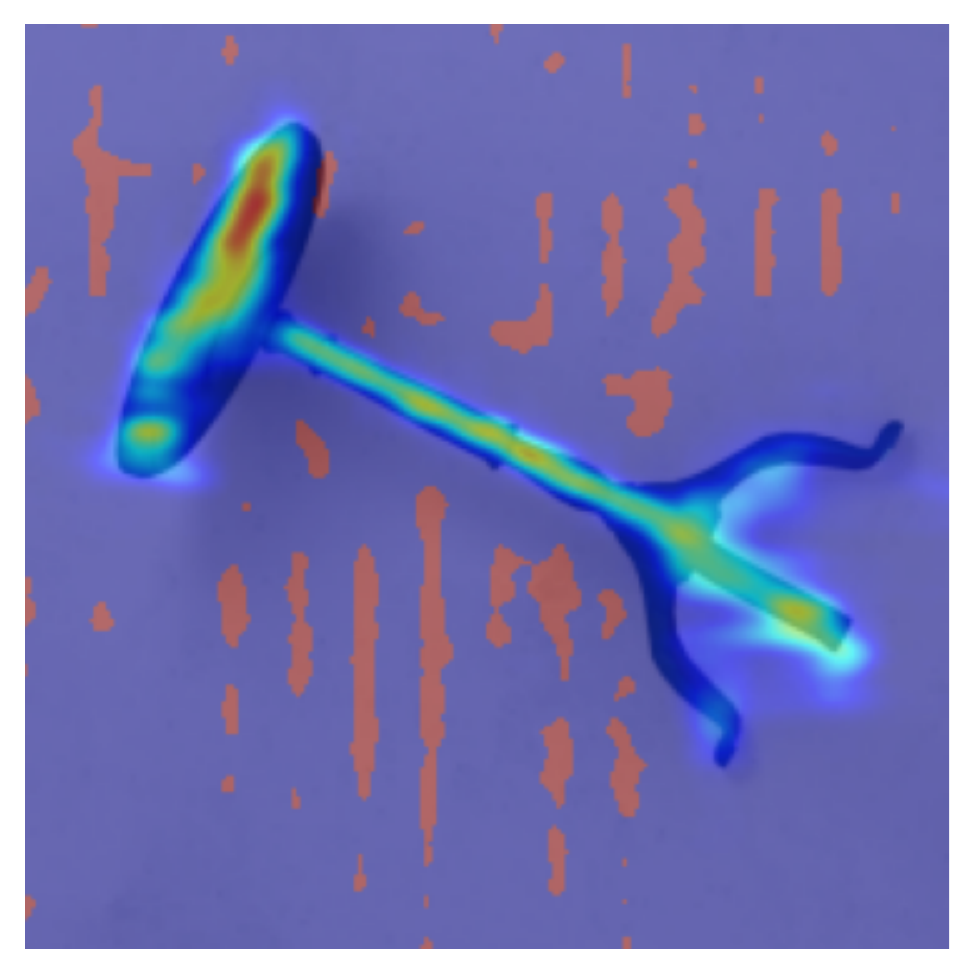}
		\includegraphics[width=0.095\textheight]{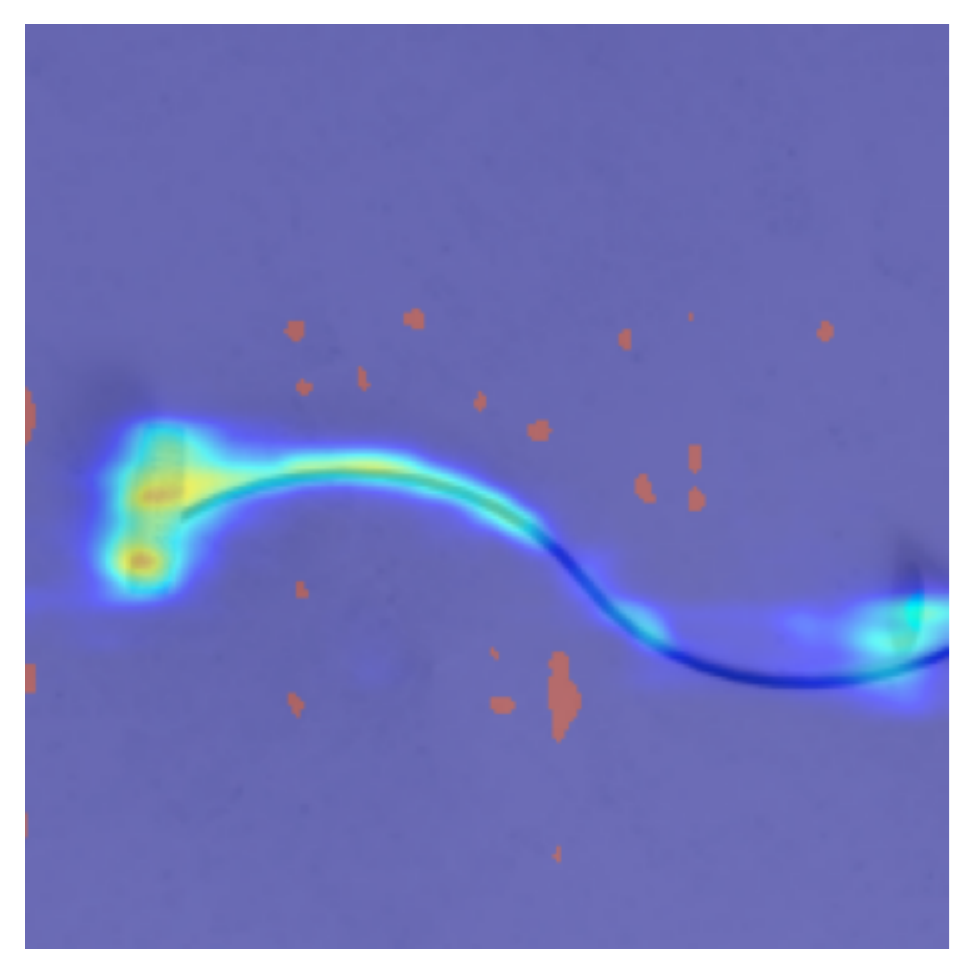}
		\includegraphics[width=0.095\textheight]{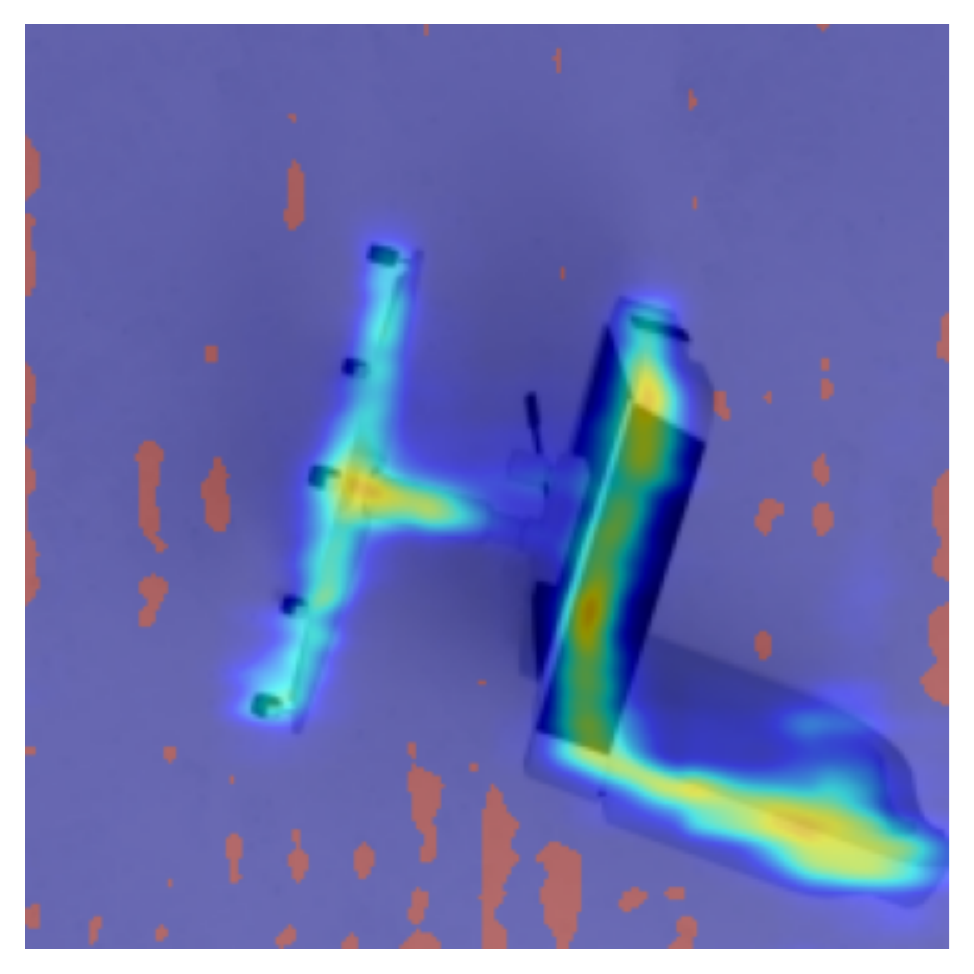}
		\includegraphics[width=0.095\textheight]{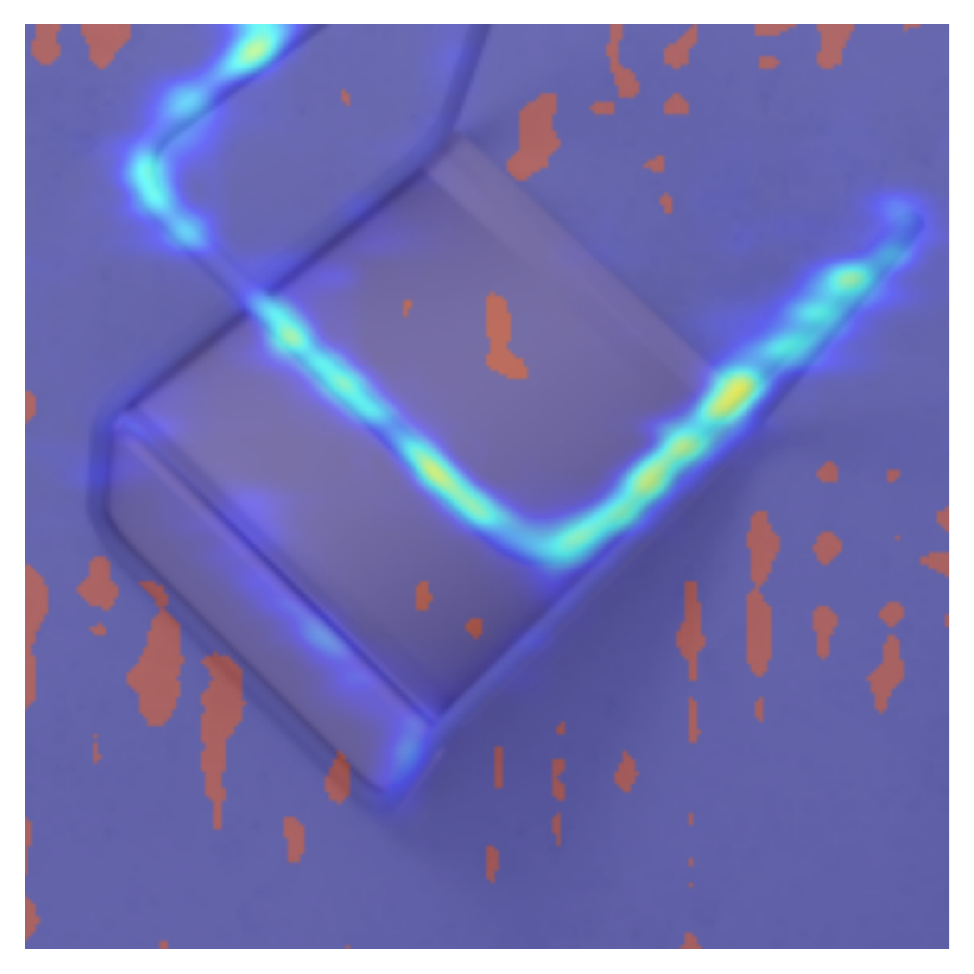}
	}
	
	\caption{The visualized attention heatmaps learned by our method, which show that our transformer model can learn the concepts beneficial for grasping.}	
	\label{attention}
\end{figure*}

\section{Experiments}
In this section,  extensive experiments are carried out to validate the performance of the proposed TF-Grasp method.  We verify the performance of TF-Grasp on two popular grasping datasets and then evaluate its effectiveness on a real Franka Panda robotic manipulator. 

The goal of this section tends to answer the following questions: 
\begin{itemize}
	\item  Is the transformer-based grasp detection model better than CNN-based models?
	\item  If true, what makes the transformer-based grasp detection model outperforming others?
\end{itemize}
\subsection{Datasets and Experiment Setup}
The Cornell grasping data \cite{lenz2015deep} is a multi-object dataset that contains 885 images.  The resolution of each image is $640 \times 480$. The whole dataset is relatively small and we use various data augmentation techniques such as rotation, zooms, and random cropping to avoid overfitting.
We then validate the performance of TF-Grasp on the Jacquard dataset\cite{depierre2018jacquard} which is generated  in a simulator via CAD models. The Jacquard dataset is fairly large, containing  over 50k images of 11k object categories, and  there are over 1 million  annotated grasp labels.

\textbf{Evaluation Metric.} A predicted grasp is  regarded as correct if the following conditions are satisfied.

\romannumeral1 ) The discrepancy between  the predicted grasping angle and the ground truth is within $30^{\circ}$.

\romannumeral2) The Jaccard index defined in Eq. \eqref{Jaccard index} is greater than $0.25$.
\begin{equation}
    \label{Jaccard index}
	J\left(\mathcal{R}^{*}, \mathcal{R}\right)=\frac{\left|\mathcal{R}^{*} \cap \mathcal{R}\right|}{\left|\mathcal{R}^{*} \cup \mathcal{R}\right|}
\end{equation}

\begin{table}[]
	\centering
		\setlength\tabcolsep{3.0pt}
	{
		\small
		\caption{The accuracy on Cornell grasping dataset. }
		\begin{tabular}{c|c|c|c|c}
			\hline   \multirow{2}{*}{\text { Method }} & \multirow{2}{*}{\text { Input }} & \multicolumn{2}{|c|}{\text { Accuracy(\%) }} & \multirow{2}{*}{\text { Time (ms) }} \\
			\cline { 3 - 4 }  & & \text { IW } & \text { OW } & \\ \hline
			
			 Fast Search \cite{jiang2011efficient}& RGB-D & 60.5 &58.3& 5000 \\ \hline
			 GG-CNN \cite{morrison2020learning} & D & 73.0&69.0 & 19   \\ \hline
			 SAE \cite{lenz2015deep} & RGB-D   & 73.9&75.6 &1350 \\ \hline
			  Two-stage closed-loop\cite{wang2016robot} & RGB-D   & 85.3&- &140 \\ \hline
			 AlexNet, MultiGrasp \cite{redmon2015real} & RGB-D &88.0 &87.1 &76 \\ \hline
			  STEM-CaRFs\cite{asif2017rgb} & RGB-D   & 88.2&87.5 & - \\ \hline
			     GRPN \cite{karaoguz2019object}& RGB   & 88.7&- & 200 \\ \hline
			  ResNet-50x2 \cite{kumra2017robotic}& RGB-D   & 89.2&88.9 & 103 \\ \hline
			  GraspNet \cite{asif2018graspnet}& RGB-D   & 90.2&90.6 & 24 \\ \hline
			  ZF-net \cite{guo2017hybrid}& RGB-D   & 93.2&89.1 & - \\ \hline
			 E2E-net \cite{ainetter2021end}  & RGB  & \textbf{98.2}&- & 63 \\ \hline
			  GR-ConvNet \cite{kumra2020antipodal} & D   & 93.2&94.3 & 19 \\ 
			  GR-ConvNet \cite{kumra2020antipodal}& RGB   & 96.6&95.5 & 19 \\ 
			   GR-ConvNet \cite{kumra2020antipodal} & RGB-D   & 97.7&96.6 & 20 \\ 	\hline	 	\hline
			      & D   & \textbf{95.2}&94.9 & 41.1 \\ 
		     TF-Grasp  & RGB     &96.78   &95.0 & 41.3 \\ 
		        & RGB-D   &\textbf{97.99}   &96.7  & 41.6 \\ 	\hline	
		\end{tabular}
		\label{cornell results}
	}
\end{table}

TF-Grasp takes a $224\times224$ image as input and outputs three pixel-wise maps with the same resolution as the input.  The input is normalized by subtracting its mean and dividing the standard  deviation.
We follow the common strategy to train the grasp transformer.  Both the encoder and decoder contain four swin-attention blocks and each consists of ${1,2,4,8}$ attention heads. The window size is 7.   At each training step, a batch of samples is randomly sampled from the training set and we use the ground truth as the target values to train our neural network. 
Concretely, we utilize the mean squared error as the loss function and apply  AdamW \cite{loshchilov2018decoupled} as the optimizer. The default size of batch size is set to $64$. The patch partition layer is implemented by convolutions with kernels of $p \times p$ and a stride $p$.
In our implementation, $p$ is set to 4.  In order to preserve a one-to-one mapping of the angle $\Theta$ between $[-\frac{\pi}{2},\frac{\pi}{2}]$, we decode the learning of angle into two components, $sin(2\Theta)$ and $cos(2\Theta)$. In this way, the final angle is obtained by $\operatorname{arctan}(\frac{sin2\Theta}{cos2\Theta})/2$.
 TF-Grasp is implemented by PyTorch, and the entire grasp detection system is running on  the Ubuntu 18.04 desktop with  Intel Core i9 CPU and NVIDIA 3090 GPU.

\subsection{Experimental Results and Analysis}
To show its effectiveness, our approach is compared with a number of  baselines under the same experimental conditions, i.e., evaluation metric. {The results of image-wise (IW) and object-wise (OW) settings in the public Cornell grasping dataset are present in Table. \ref{cornell results}. Since the Cornell dataset is relatively small, we follow the setting of previous works \cite{kumra2017robotic},\cite{redmon2015real},\cite{lenz2015deep} by adopting a five-fold cross-validation. Also, to make the comparison fair and comprehensive, the input modalities and running time are considered. For all compared baselines, we use the data reported in their original papers.}  Taking as input only the depth information, our TF-Grasp achieves an accuracy of $95.2 \%$ which is competitive to the state-of-the-arts. When using both depth and RGB data, our model obtains $97.99\%$ accuracy. { For Table \ref{jcaquard}, we use $90\%$ data of the Jacquard dataset as the  training set and the remaining $10\%$ as the validation set. In addition, our model takes about $41$ms to process a single image using the Intel Core i9-10900X CPU processor, which is competitive with the state-of-art approaches and basically meets the real-time requirements.}
The transformer grasping model exhibits a better accuracy on both datasets compared to conventional CNN models.  Our proposed approach achieves a higher accuracy of $94.6\%$ which is on-par or superior to previous methods. The results  on the Cornell and Jacquard datasets all indicate that the model with the attention mechanism is more suitable for visual grasping tasks.

\begin{figure}[H]
	\center
	\subfigure{
		\includegraphics[width=0.26\textheight]{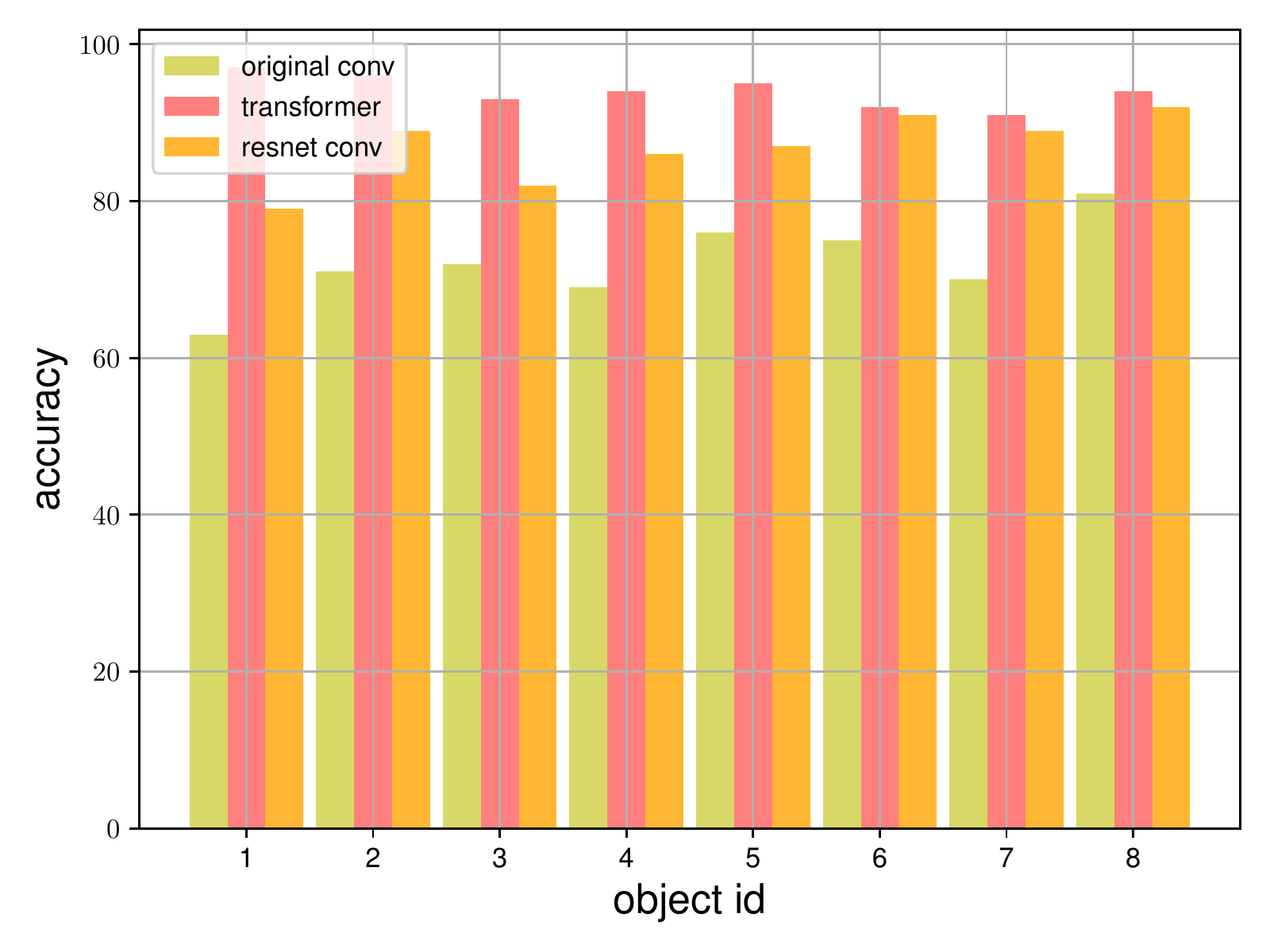}
	}
	\caption{ The accuracy of different models as feature extractors on selected objects. }	
	\label{comp}
\end{figure}

\begin{table}[]
	\begin{center}
		\caption{The accuracy on Jacquard grasping dataset. } 
		\begin{tabular}{l|l|l|c}
			
			\hline Authors &Method & Input  & Accuracy $(\%)$ \\
			\hline Depierre \cite{depierre2018jacquard} &Jacquard& RGB-D & $74.2$ \\
			Morrison \cite{morrison2020learning}  & GG-CNN2 &D     & 84 \\
			Zhou \cite{zhou2018fully} & FCGN, ResNet-101& RGB & $91.8$ \\
			Alexandre \cite{gariepy2019gq} & GQ-STN &D &$70.8$ \\
			Zhang \cite{zhang2019roi}& ROI-GD &RGB &$90.4$ \\
			
			Stefan \cite{ainetter2021end} &Det Seg &RGB&  $92.59$\\
			Stefan \cite{ainetter2021end} &Det Seg Refine &RGB&  $92.95$\\
			Kumra \cite{kumra2020antipodal}     &   GR-ConvNet & D   & $\mathbf{93.7}$ \\
			Kumra  \cite{kumra2020antipodal}    &   GR-ConvNet & RGB  & 91.8 \\
			Kumra  \cite{kumra2020antipodal}    &   GR-ConvNet & RGB-D   & $\mathbf{94.6}$ \\
			\hline & TF-Grasp&  D & $93.1$ \\
			Our & TF-Grasp  &RGB & $\mathbf{93.57}$ \\
			& TF-Grasp  &RGB-D & $\mathbf{9 4 . 6}$ \\
			
			\hline
		\end{tabular}
		\label{jcaquard}
	\end{center}
\end{table}
Despite the fact that our model is trained on a single object dataset,
it can be well adapted to multi-object environments with the help of attention mechanisms.
In addition, to evaluate the advantages of the transformer versus CNNs for visual grasping tasks, we use the original convolution layers, residual layers, and our transformer as feature extractors to test detection accuracy on different objects on the Cornell dataset. {We apply an object-wise split to the Cornell dataset and Fig. \ref{comp} shows the detection accuracy of objects not seen during the training phase.  All objects are subsets of the Cornell dataset and are evaluated 5 times. All models shown in Fig. \ref{comp} employ an encoder-decoder architecture with 4 stages in order to guarantee a fair comparison, where the original-conv is a fully convolutional neural network and resnet-conv is to replace the original convolution layer  with the residual block. }
The result of different models is shown in Fig. \ref{comp}. Note that the transformer outperforms original convolutions on all selected objects and is marginally better or on-par with the residual network.

These results demonstrate that the transformer improves robotic grasp detection.  We conjecture that prior methods that rely on local operations of the convolution layers might ignore the dependencies between long-range pixels. Instead, our approach leverages the attention mechanism to exploit both local and global information and integrates features that are useful for grasping.
\begin{figure*}[]
		\center
		
	\subfigure[ Samples of generated rectangles predicted by CNN]{
		
		\includegraphics[width=0.077\textheight]{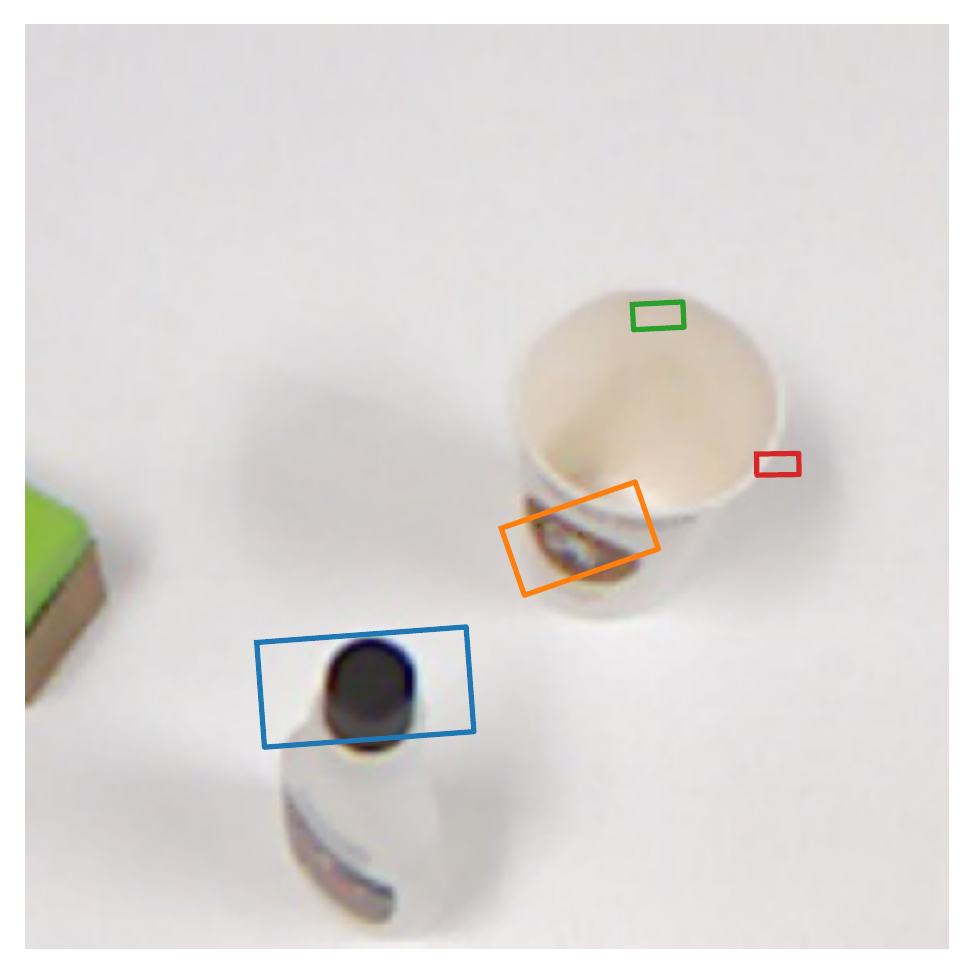}
		\includegraphics[width=0.077\textheight]{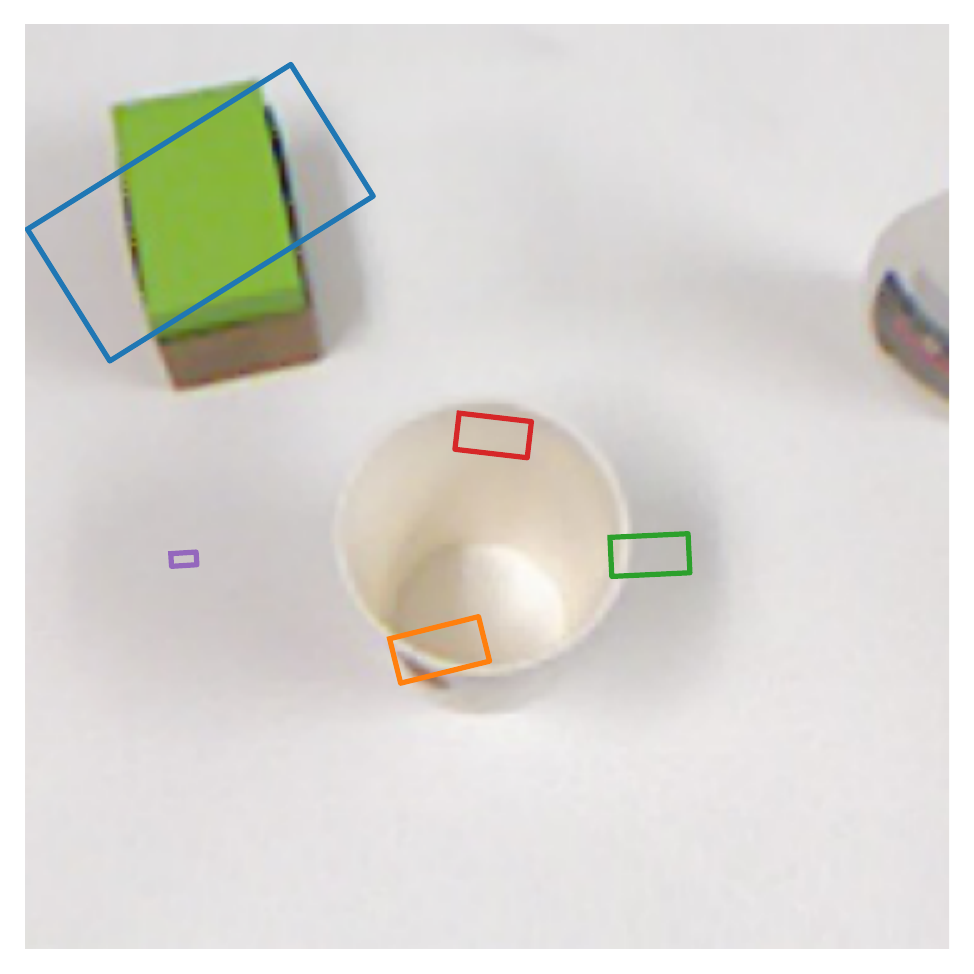}
		\includegraphics[width=0.077\textheight]{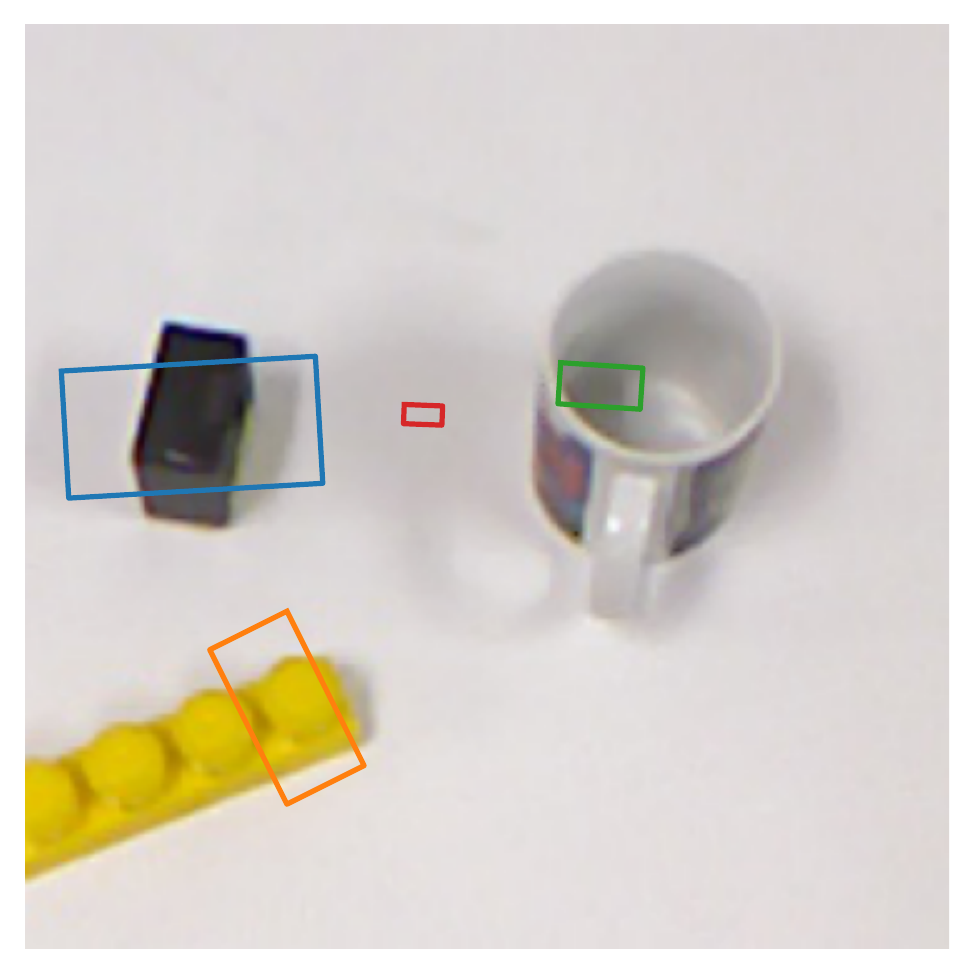}
		\includegraphics[width=0.077\textheight]{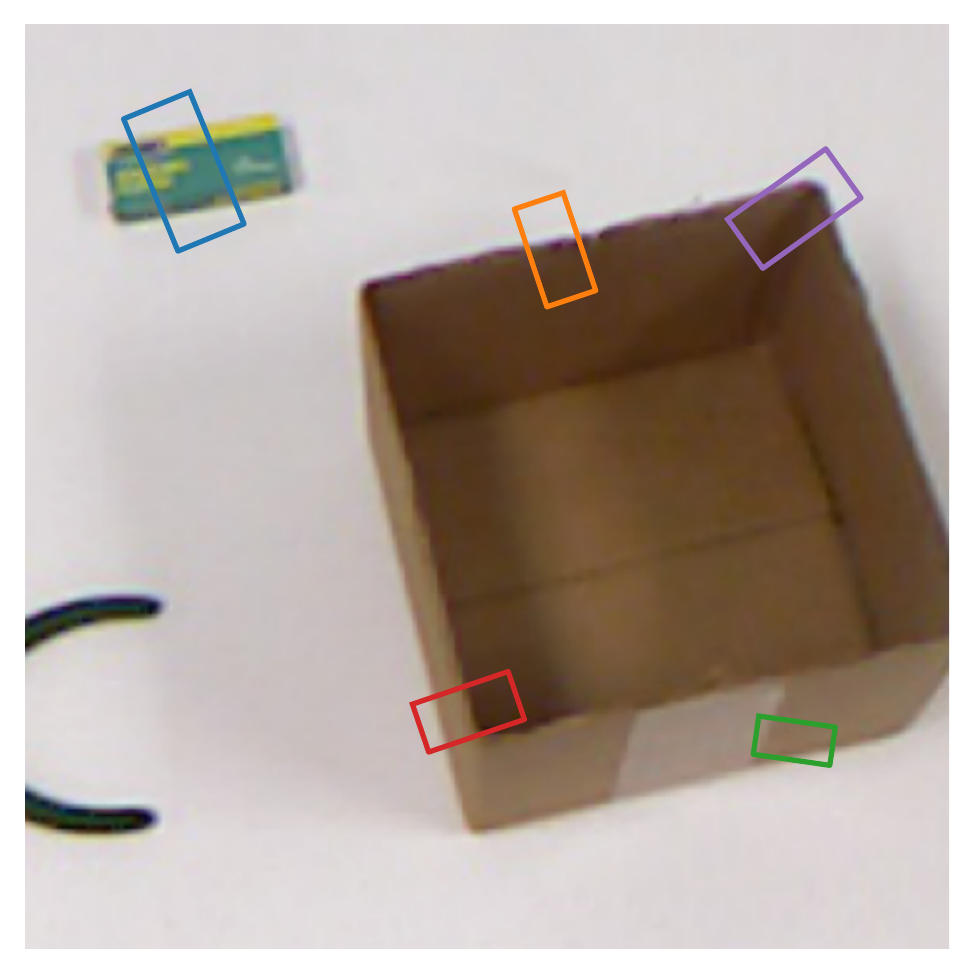}
		\includegraphics[width=0.077\textheight]{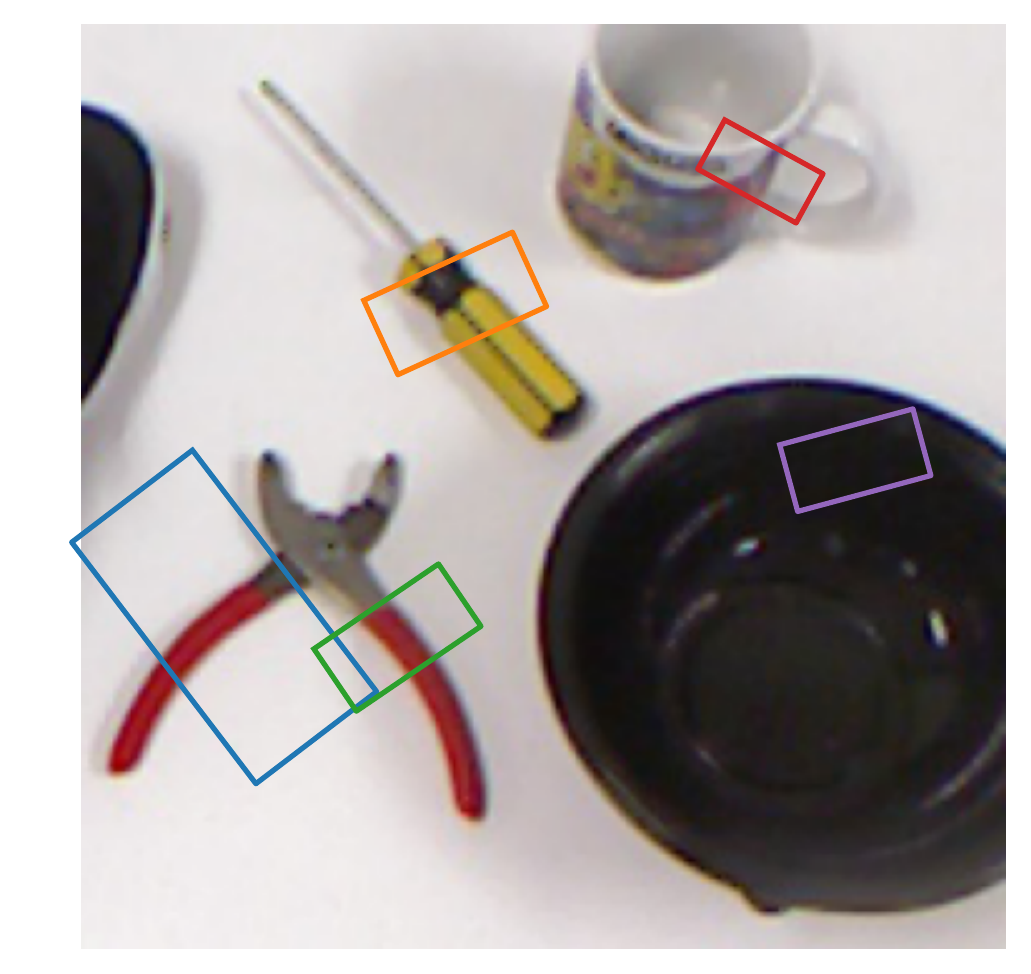}
		\includegraphics[width=0.077\textheight]{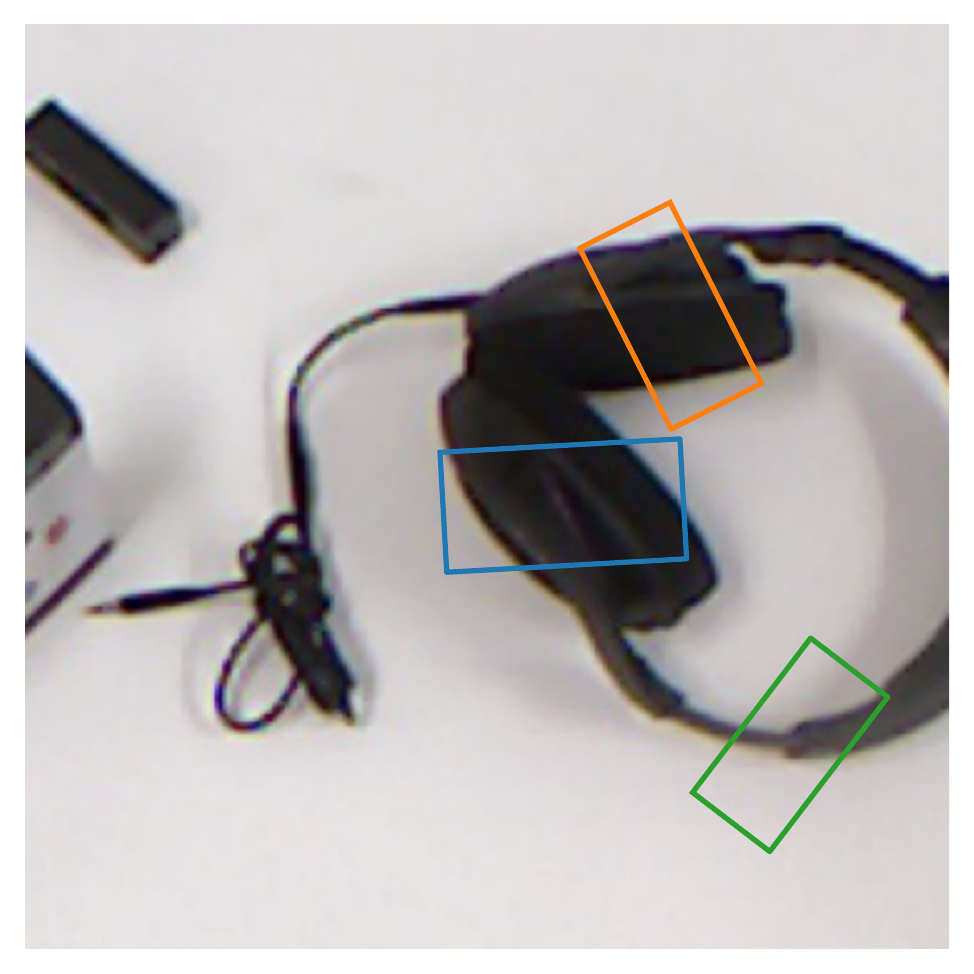}
		\includegraphics[width=0.077\textheight]{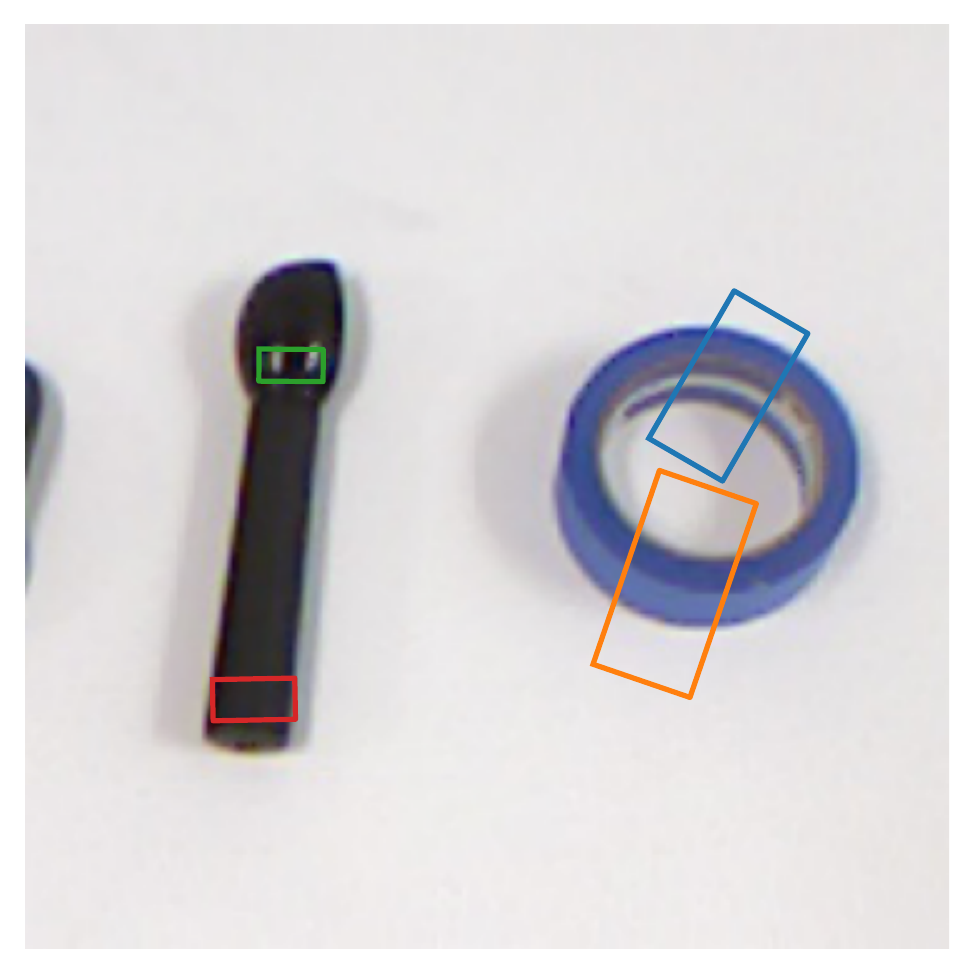}
		\includegraphics[width=0.077\textheight]{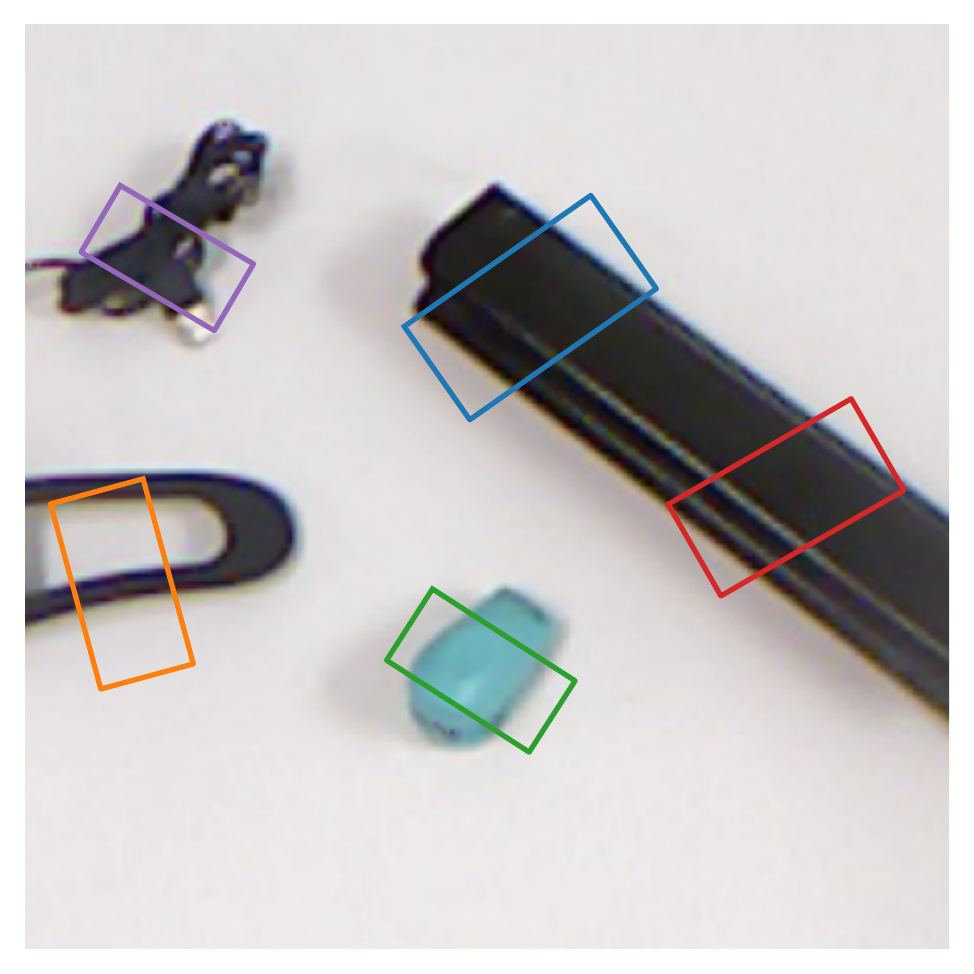}
		
	}
	\subfigure[Predicted grasp  quality heatmaps by CNN]{	
		\includegraphics[width=0.077\textheight]{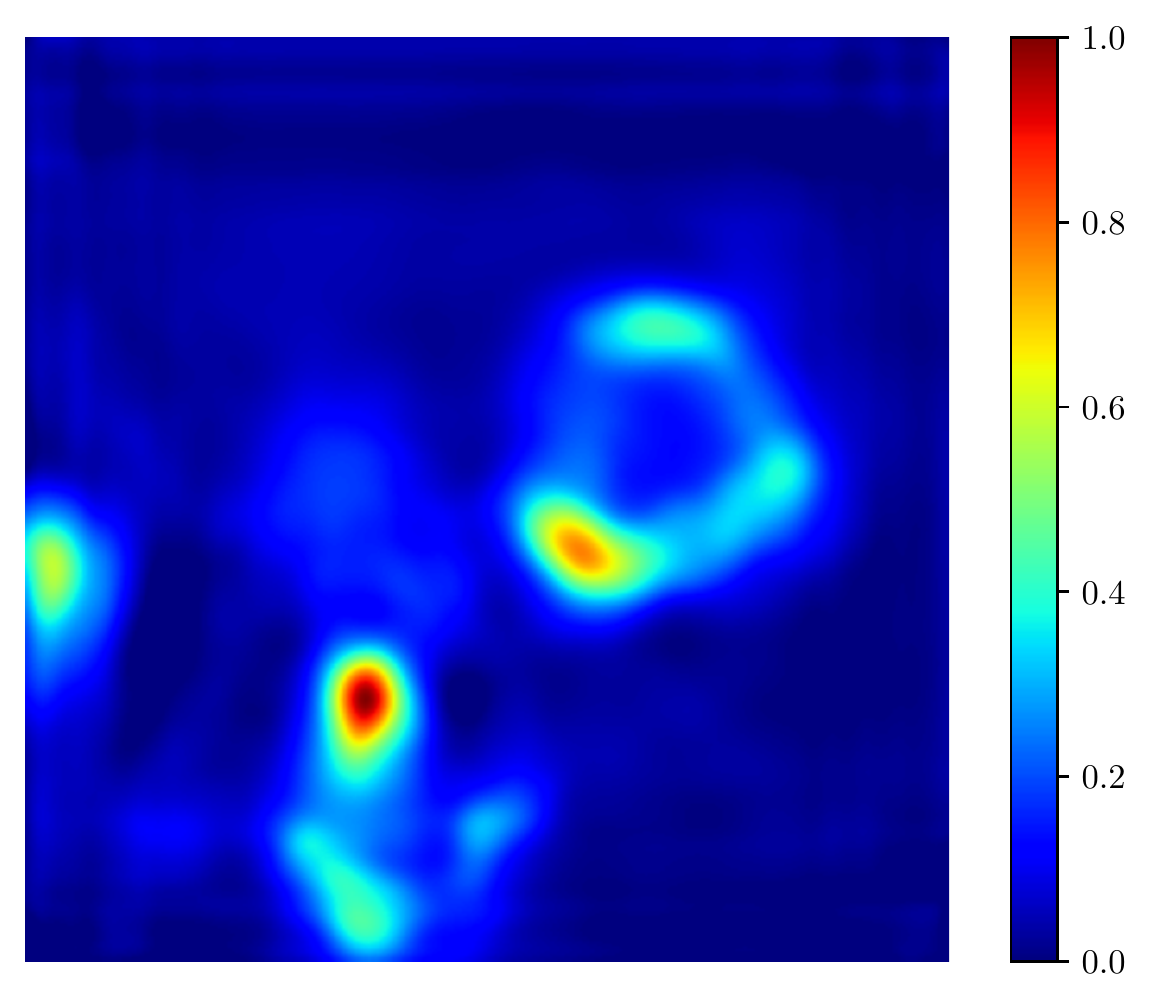}
		\includegraphics[width=0.077\textheight]{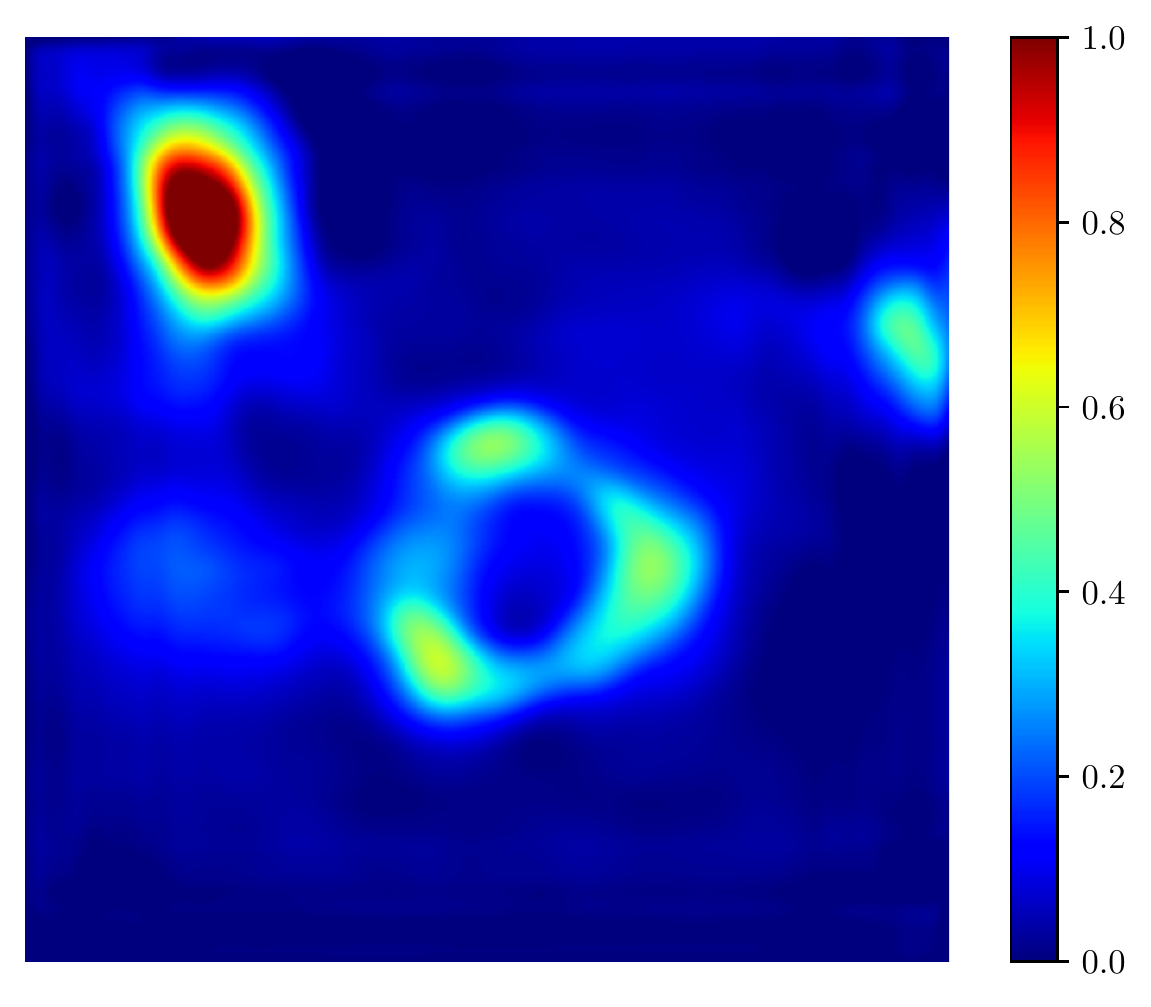}
		\includegraphics[width=0.077\textheight]{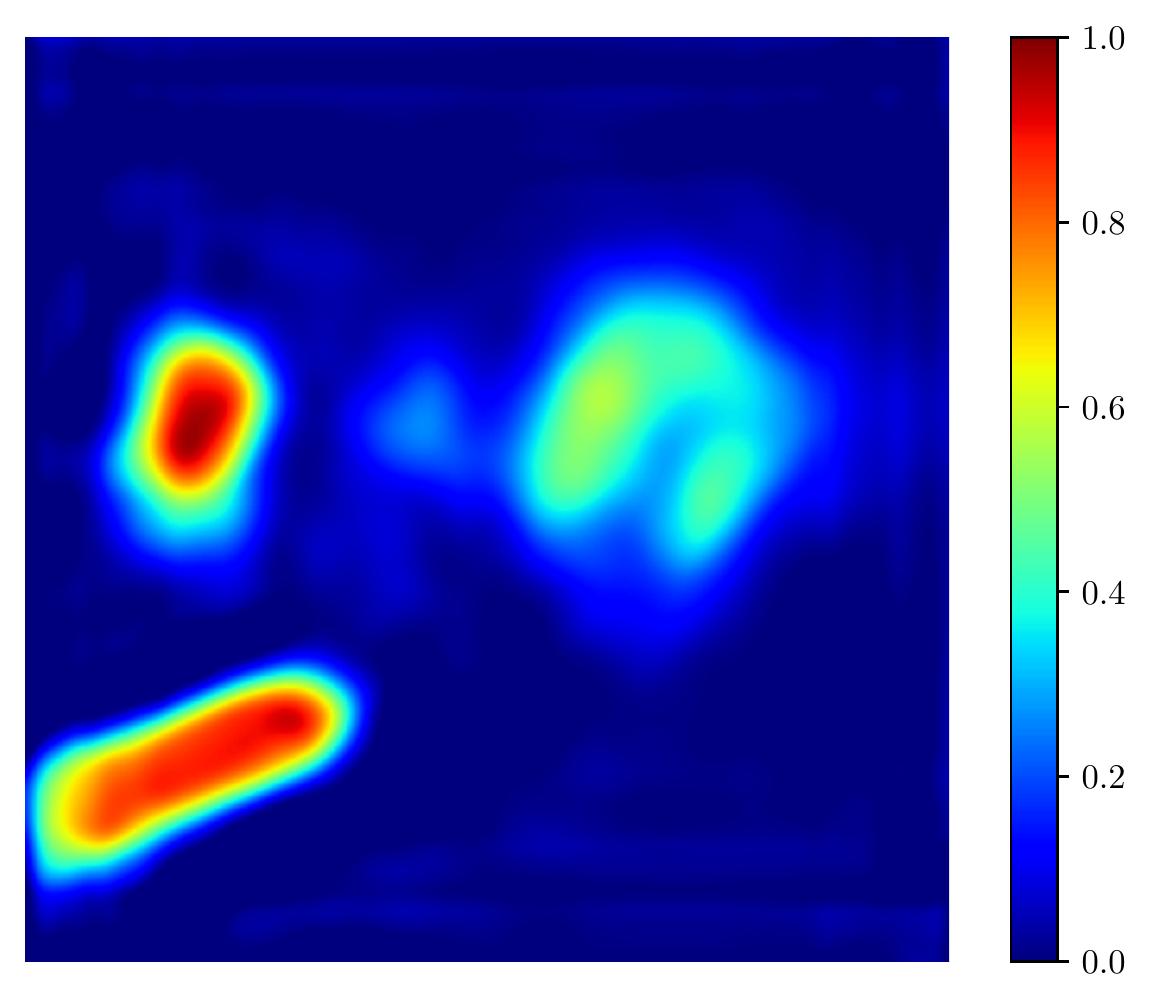}
		\includegraphics[width=0.077\textheight]{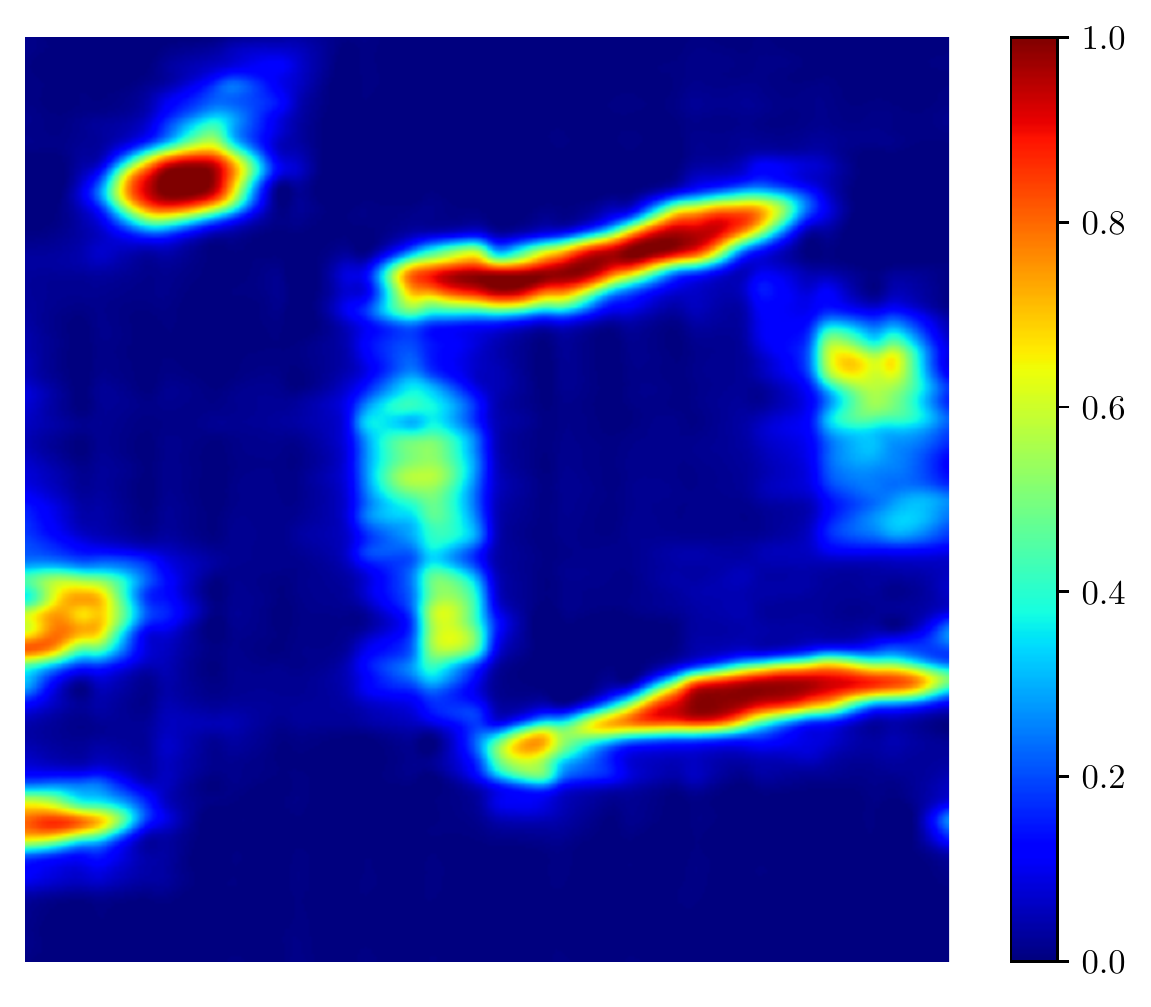}
		\includegraphics[width=0.077\textheight]{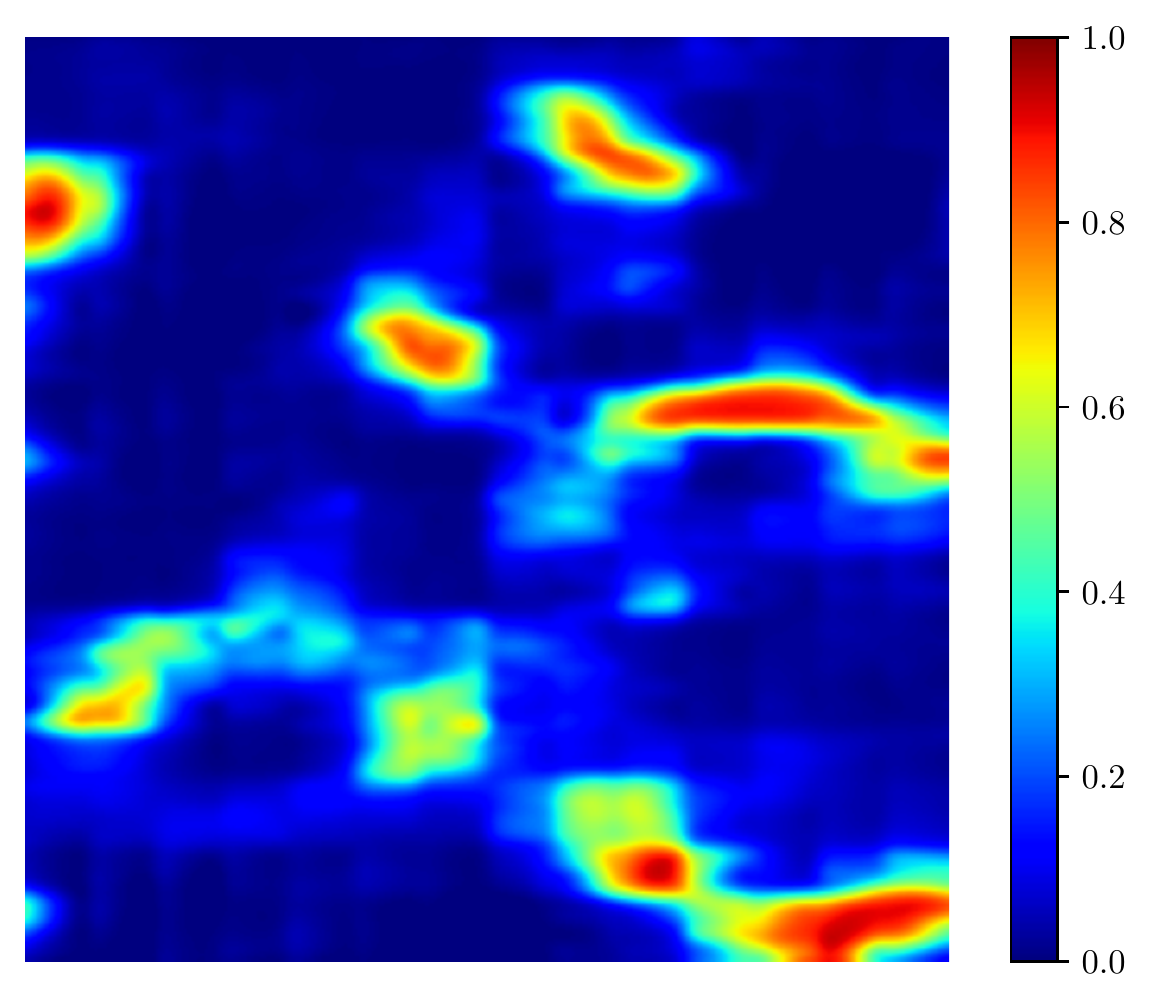}
		\includegraphics[width=0.077\textheight]{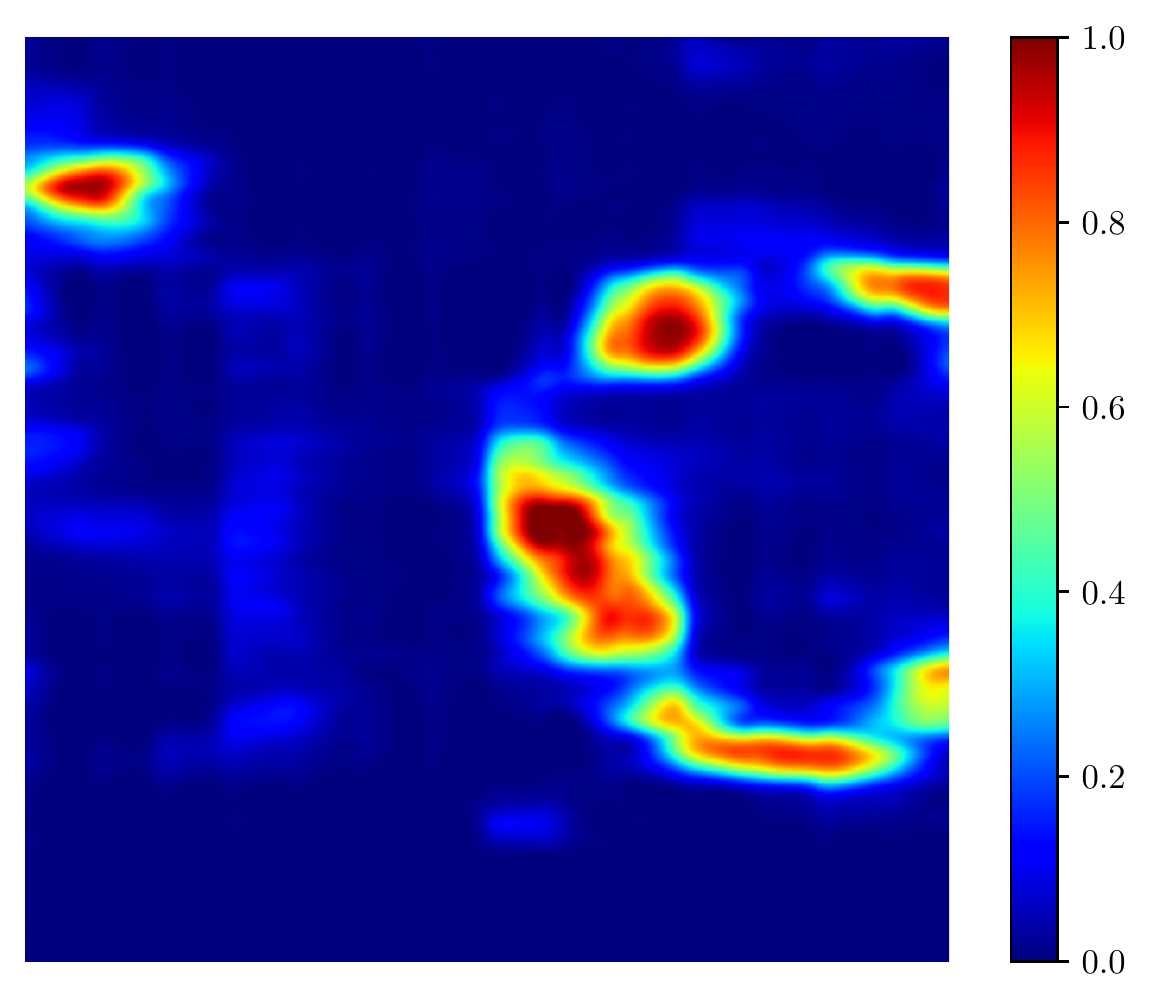}
		\includegraphics[width=0.077\textheight]{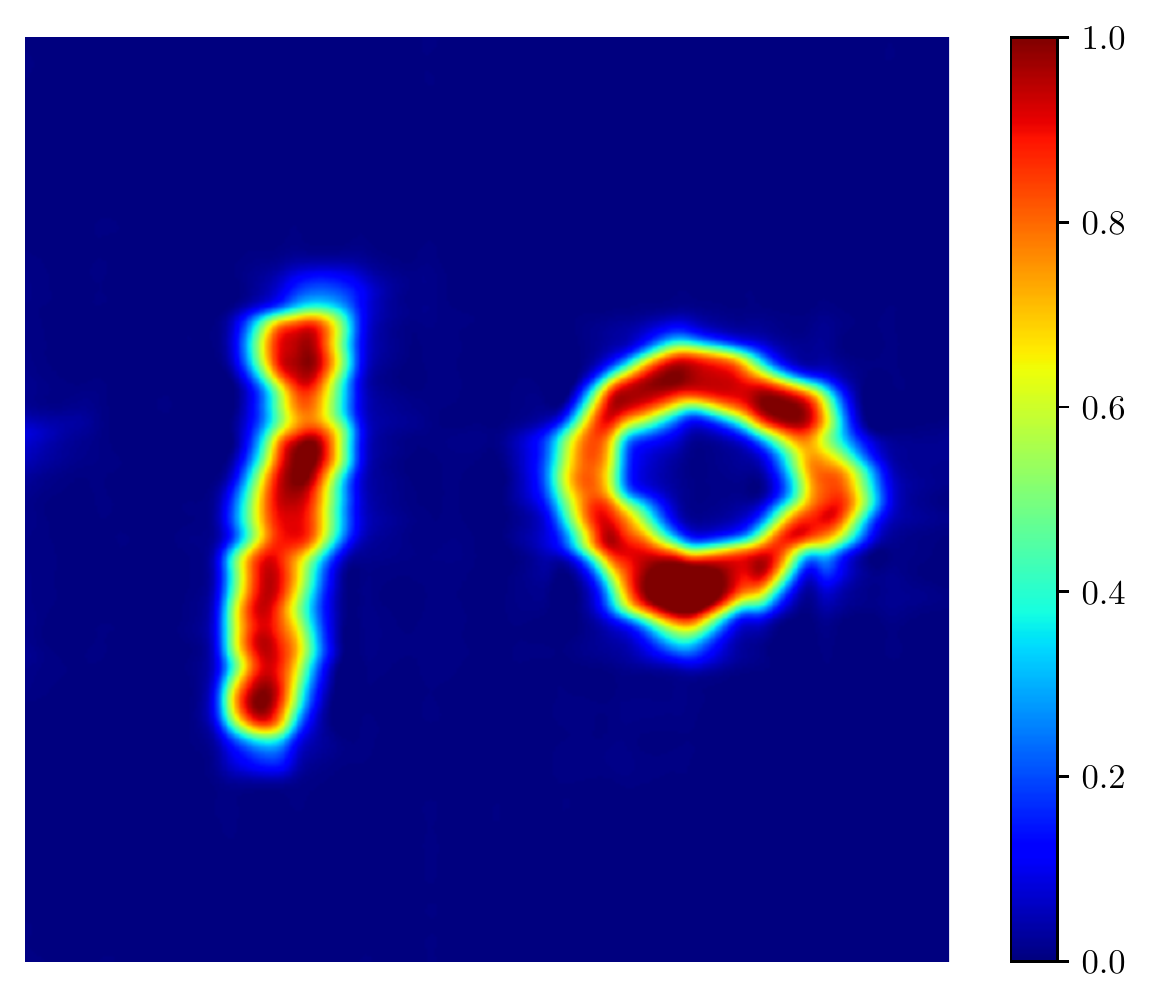}
		\includegraphics[width=0.077\textheight]{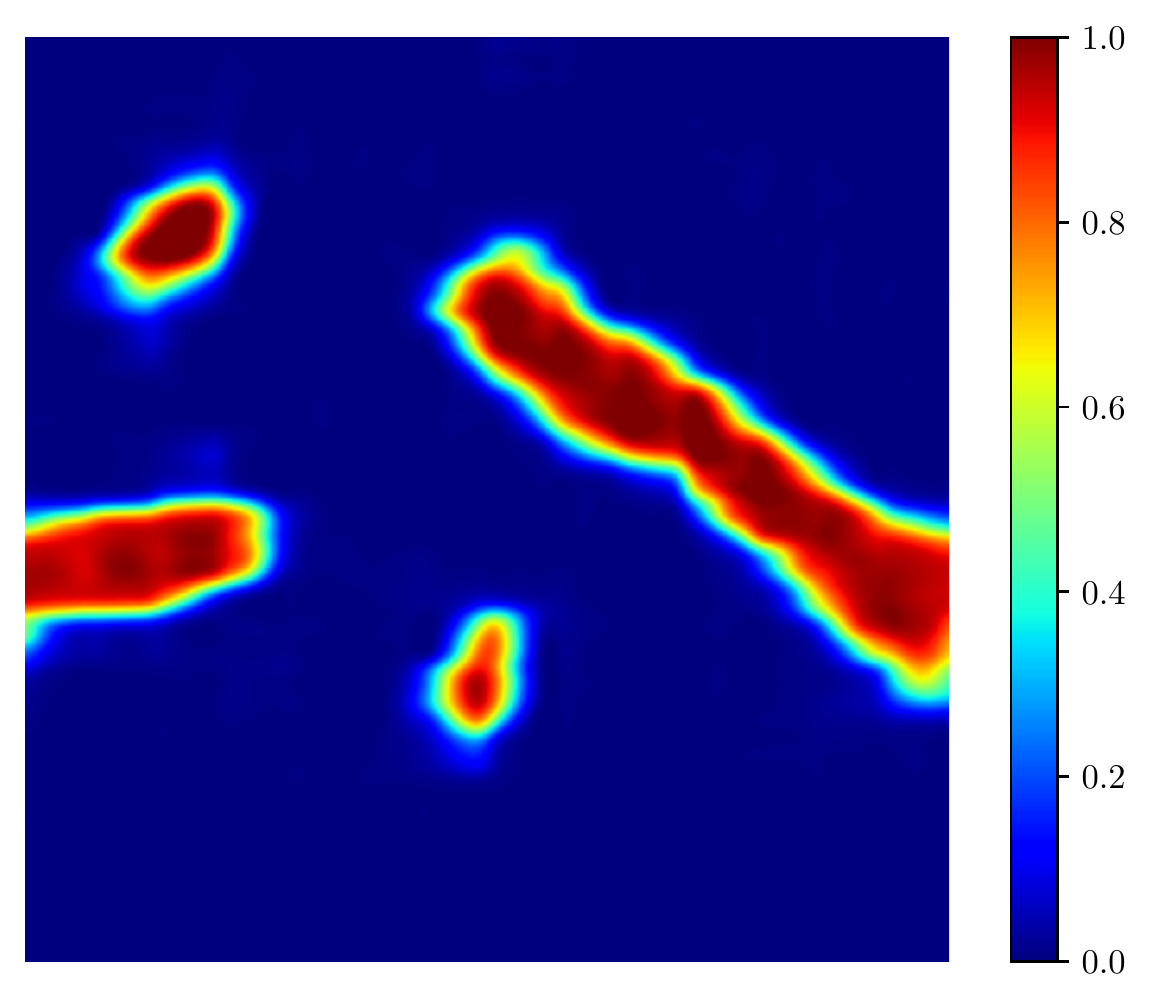}
	}
	
	\subfigure[Samples of generated rectangles predicted by Transformer]{	
		\includegraphics[width=0.077\textheight]{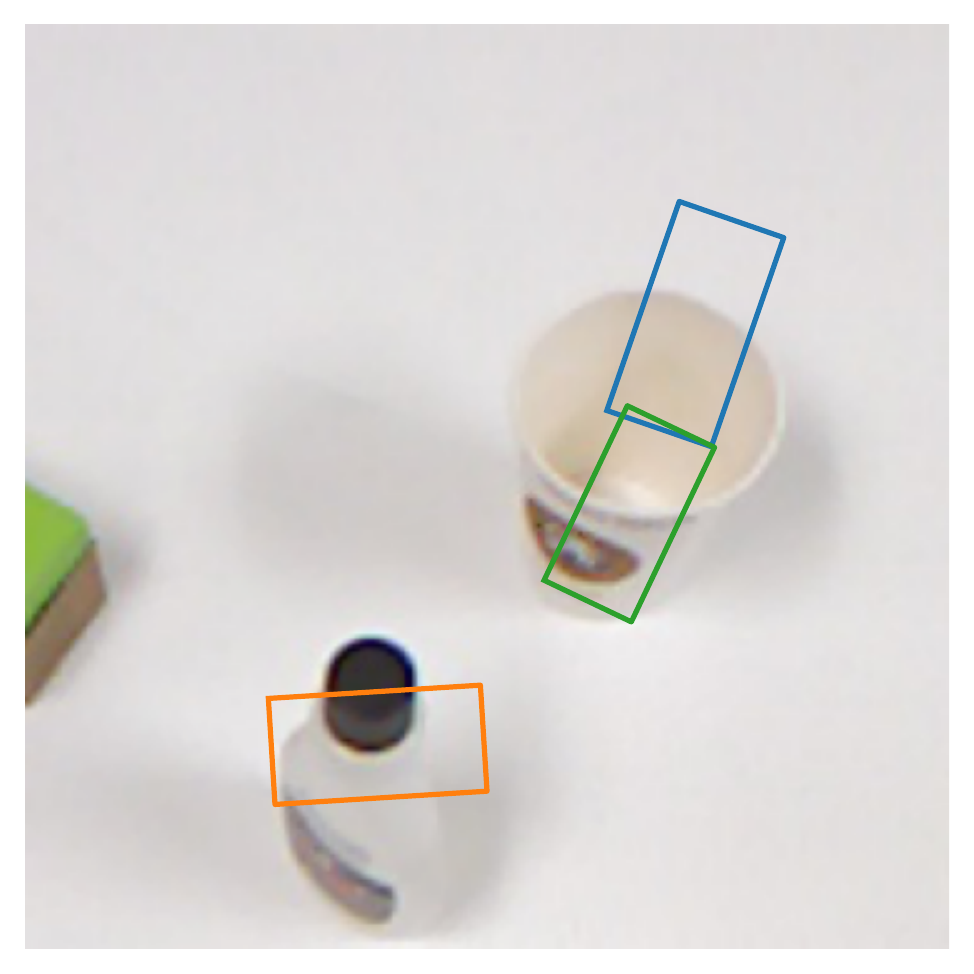}
		\includegraphics[width=0.077\textheight]{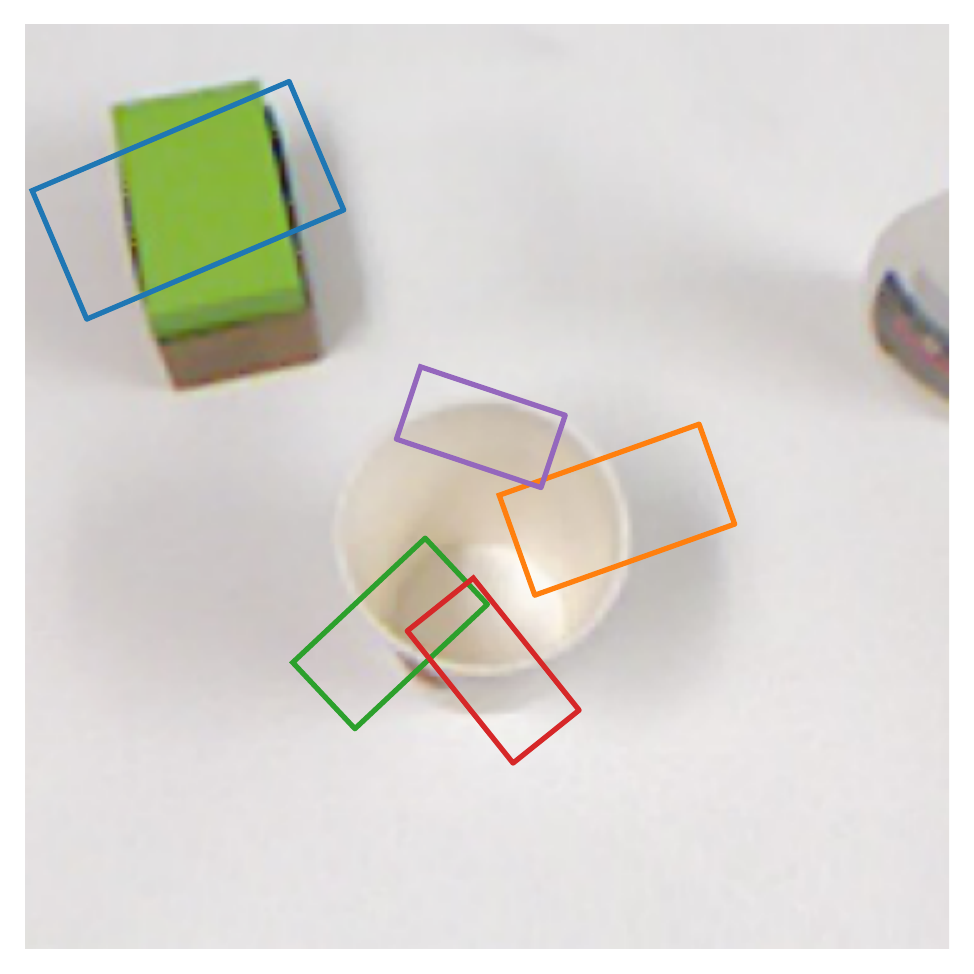}
		\includegraphics[width=0.077\textheight]{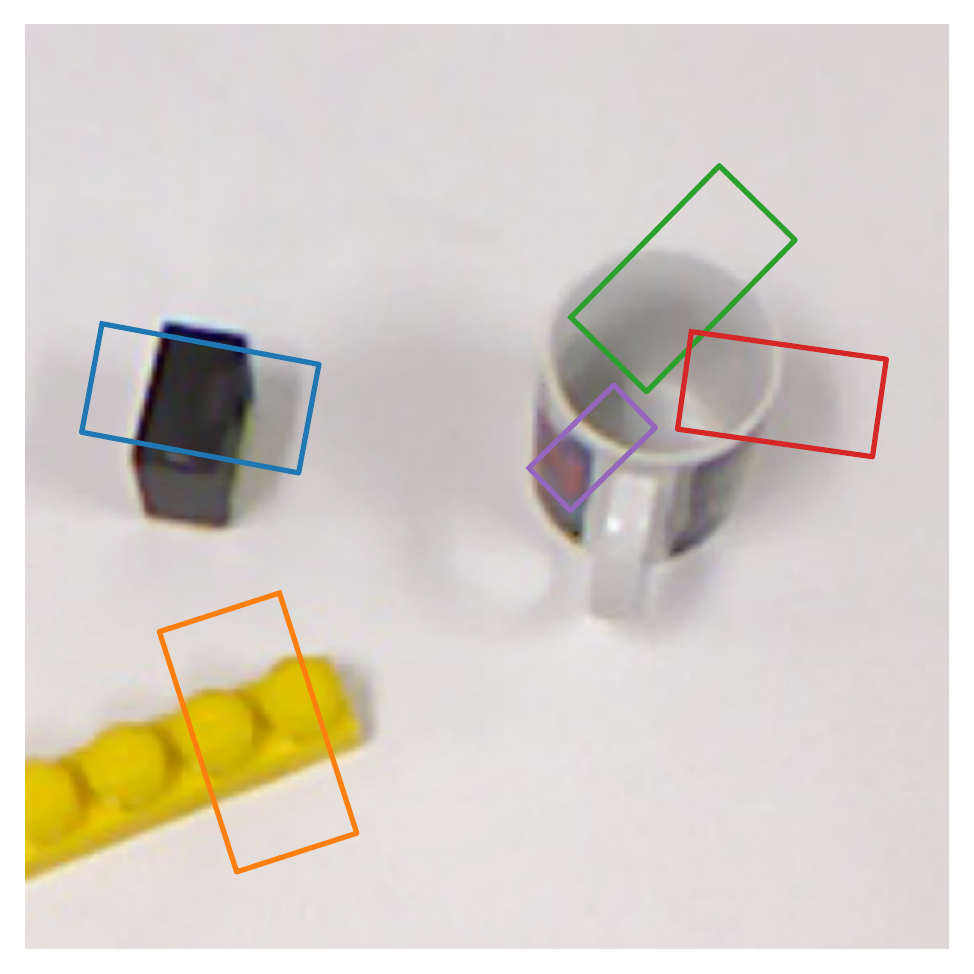}
		\includegraphics[width=0.077\textheight]{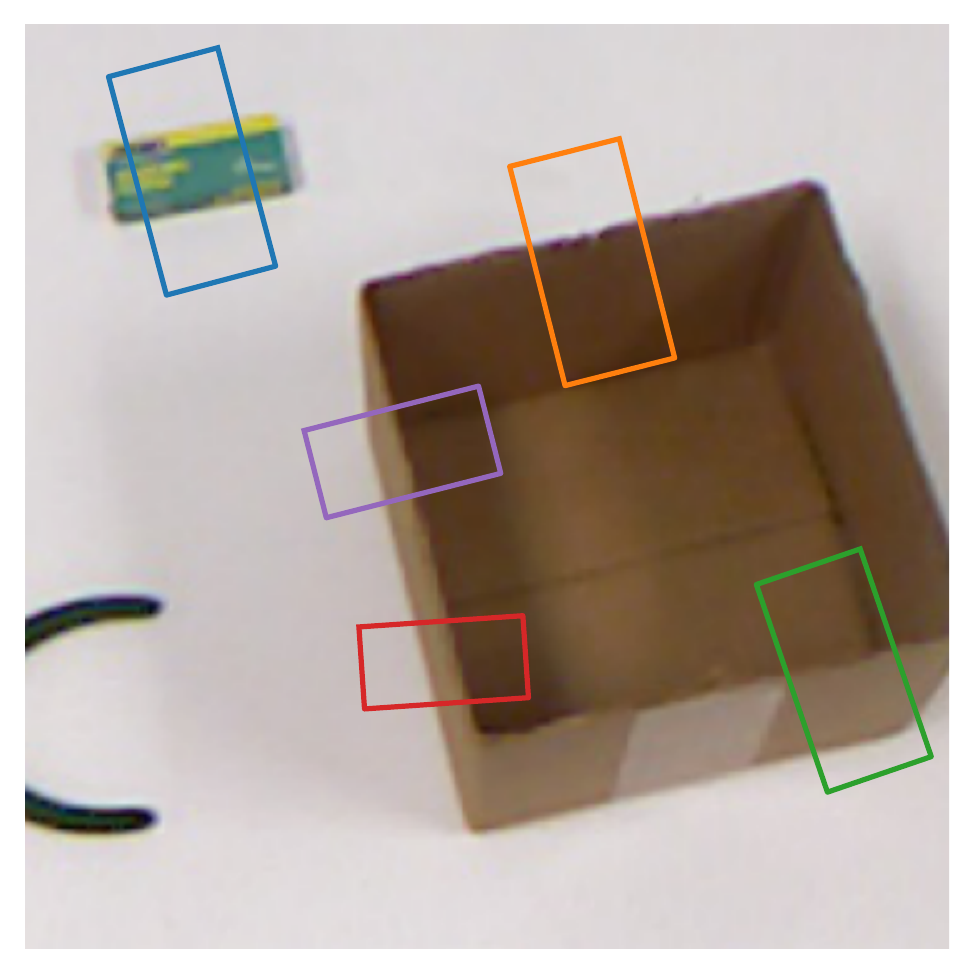}
		\includegraphics[width=0.077\textheight]{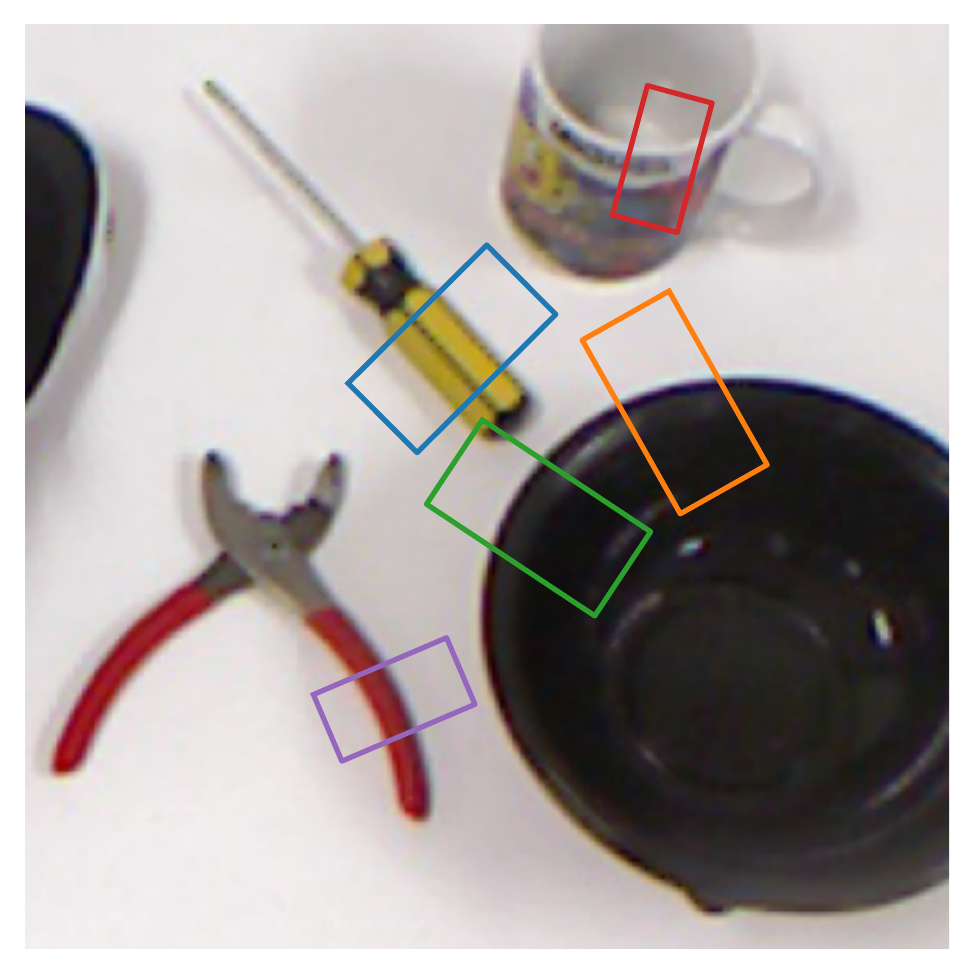}
		\includegraphics[width=0.077\textheight]{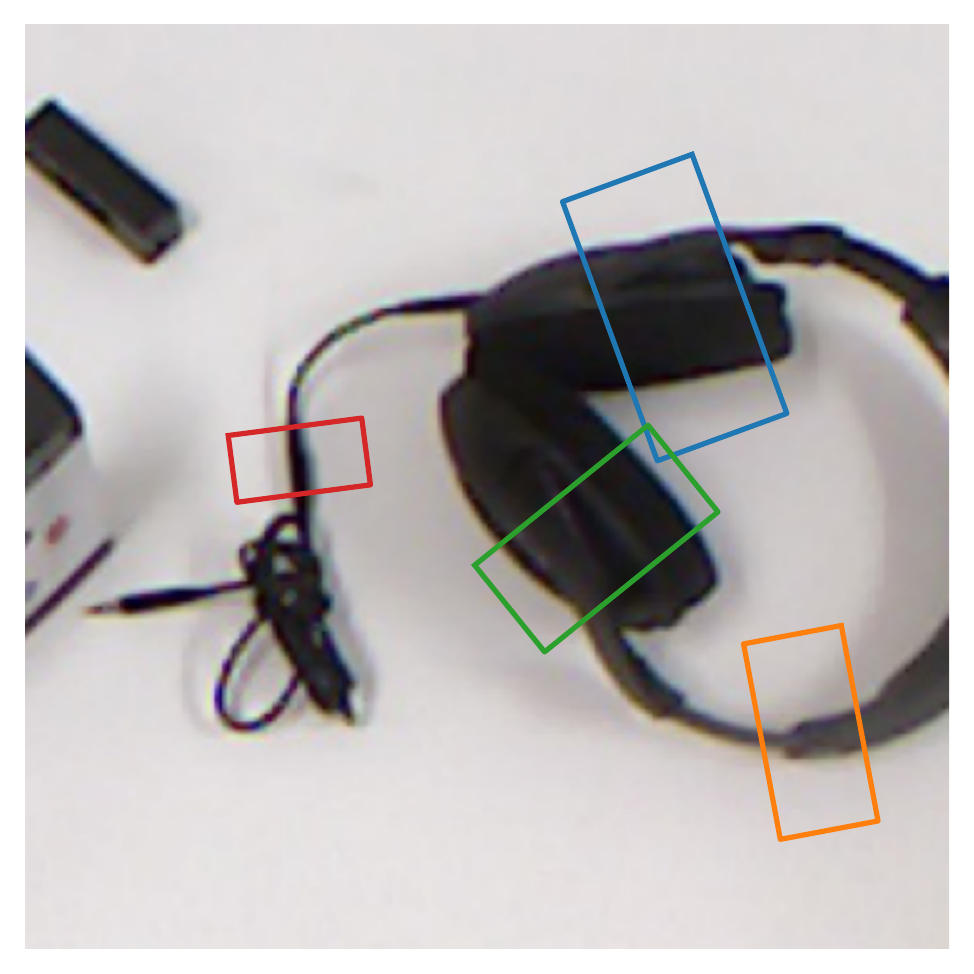}
		\includegraphics[width=0.077\textheight]{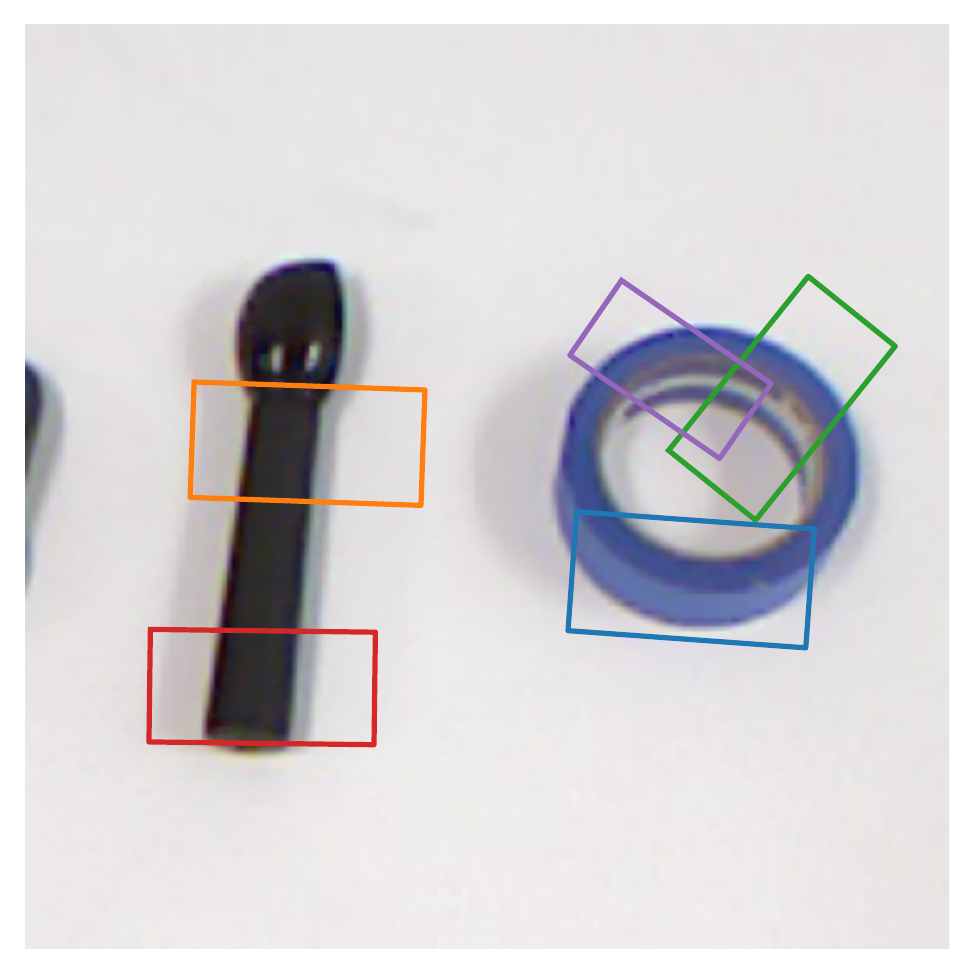}
		\includegraphics[width=0.077\textheight]{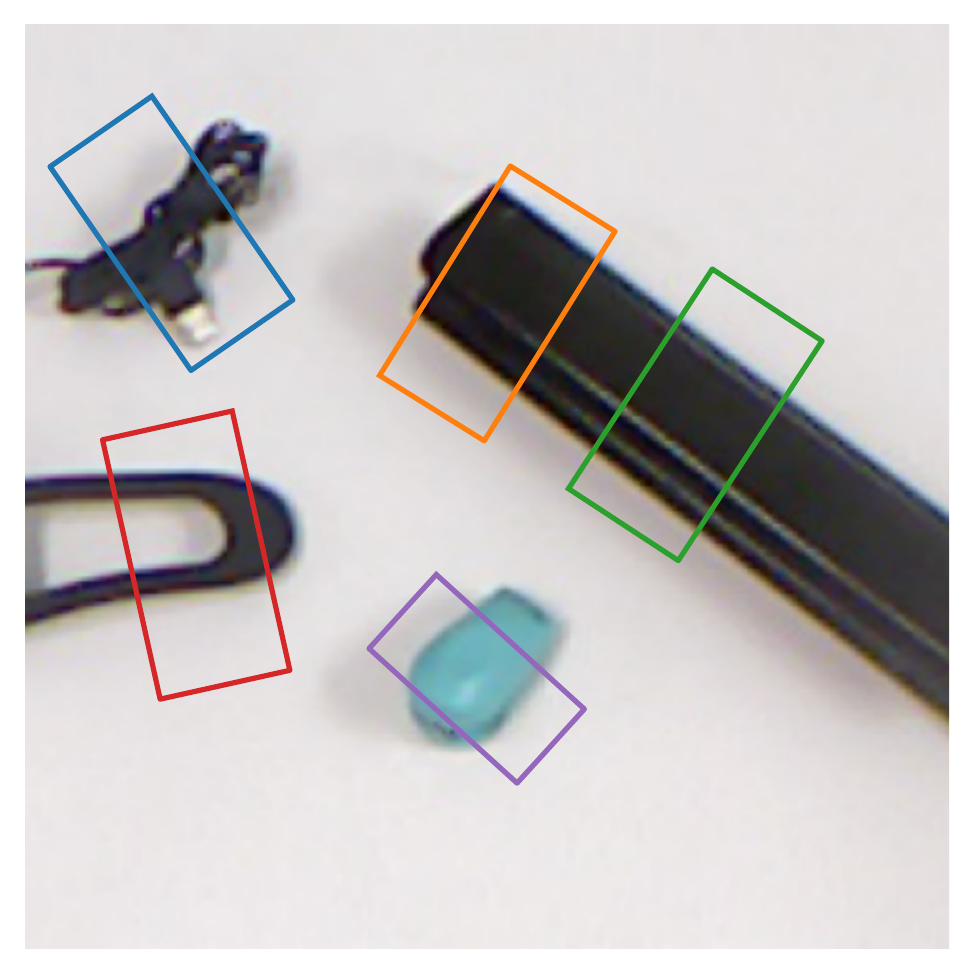}
	}
	\subfigure[Predicted grasp quality heatmaps by  Transformer]{	
		\includegraphics[width=0.077\textheight]{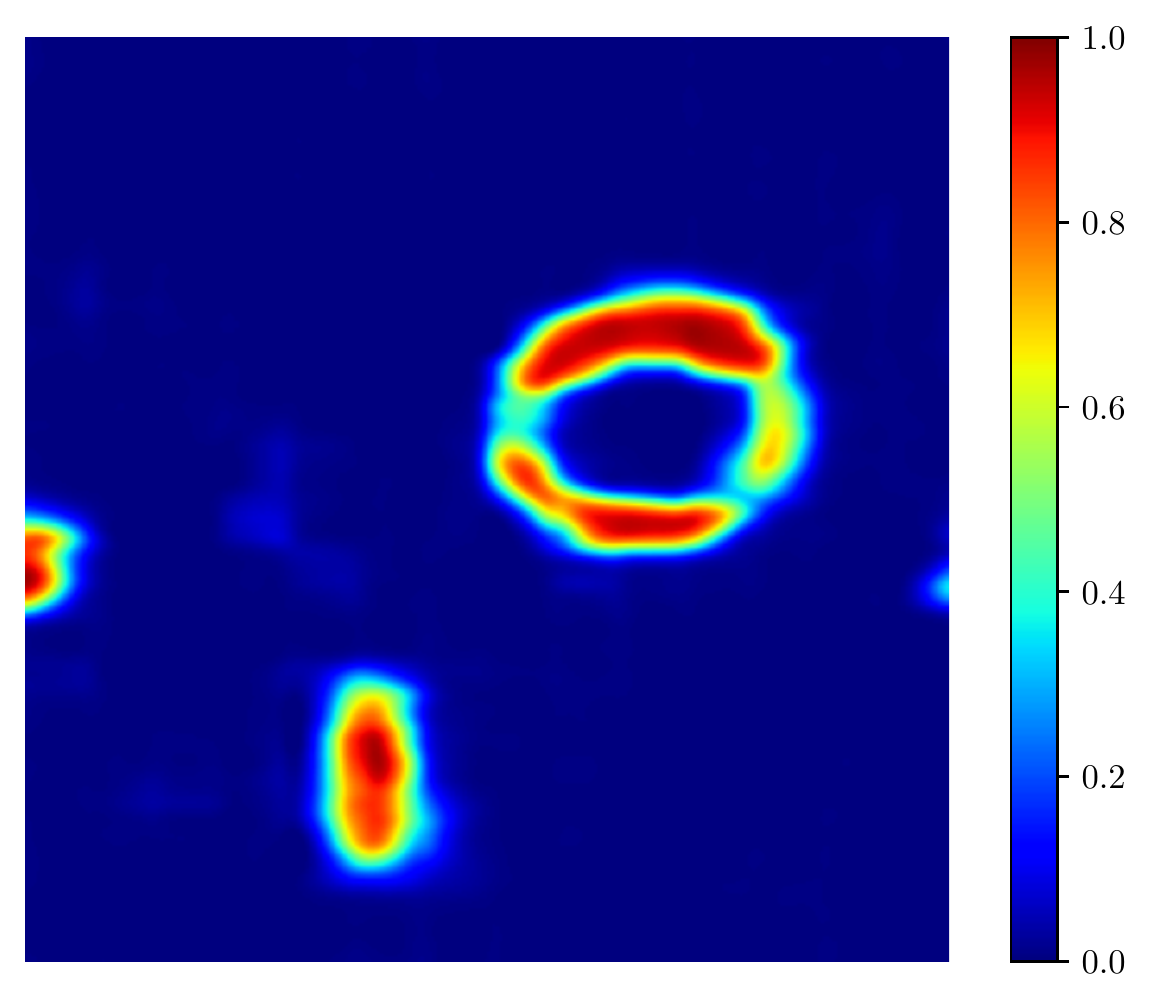}
		\includegraphics[width=0.077\textheight]{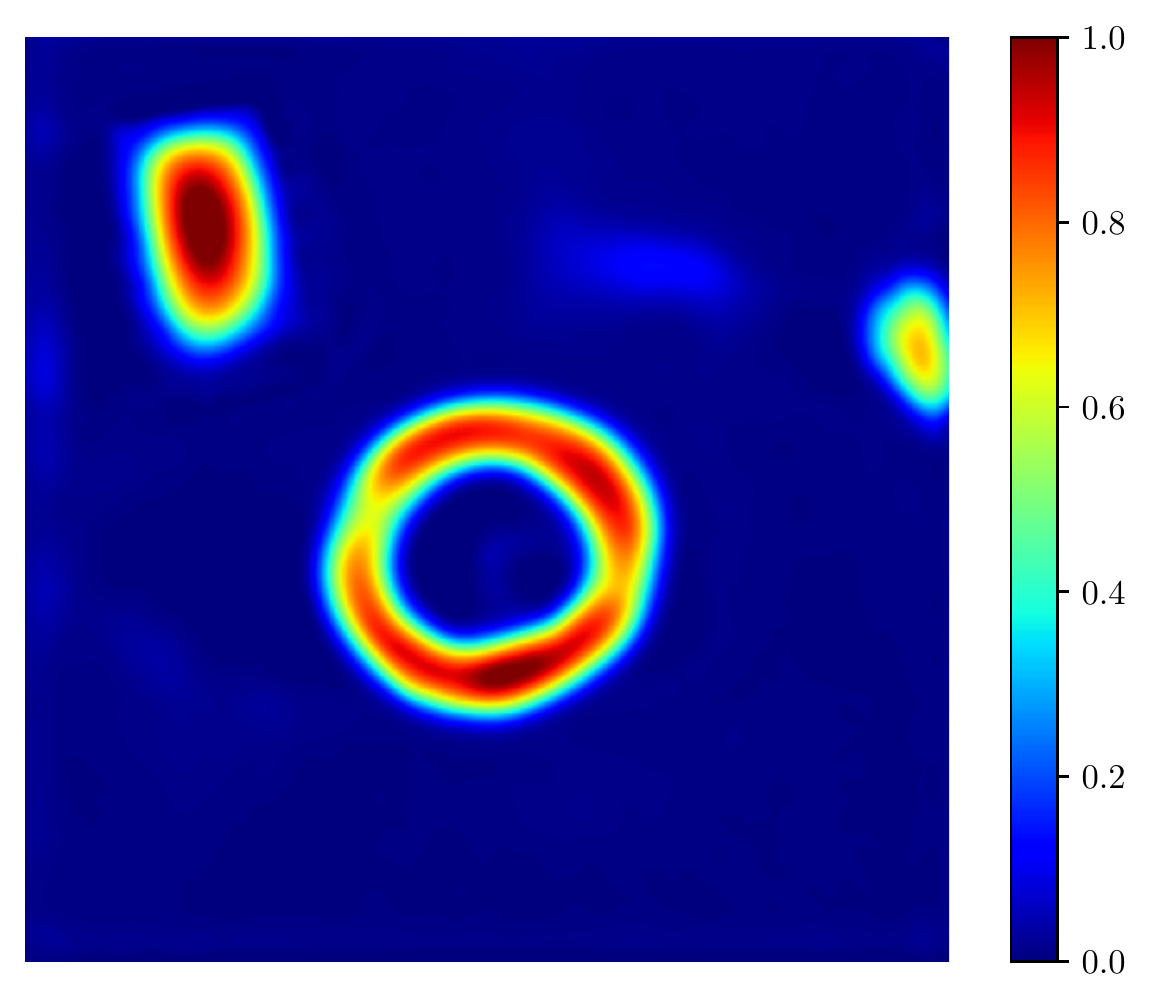}
		\includegraphics[width=0.077\textheight]{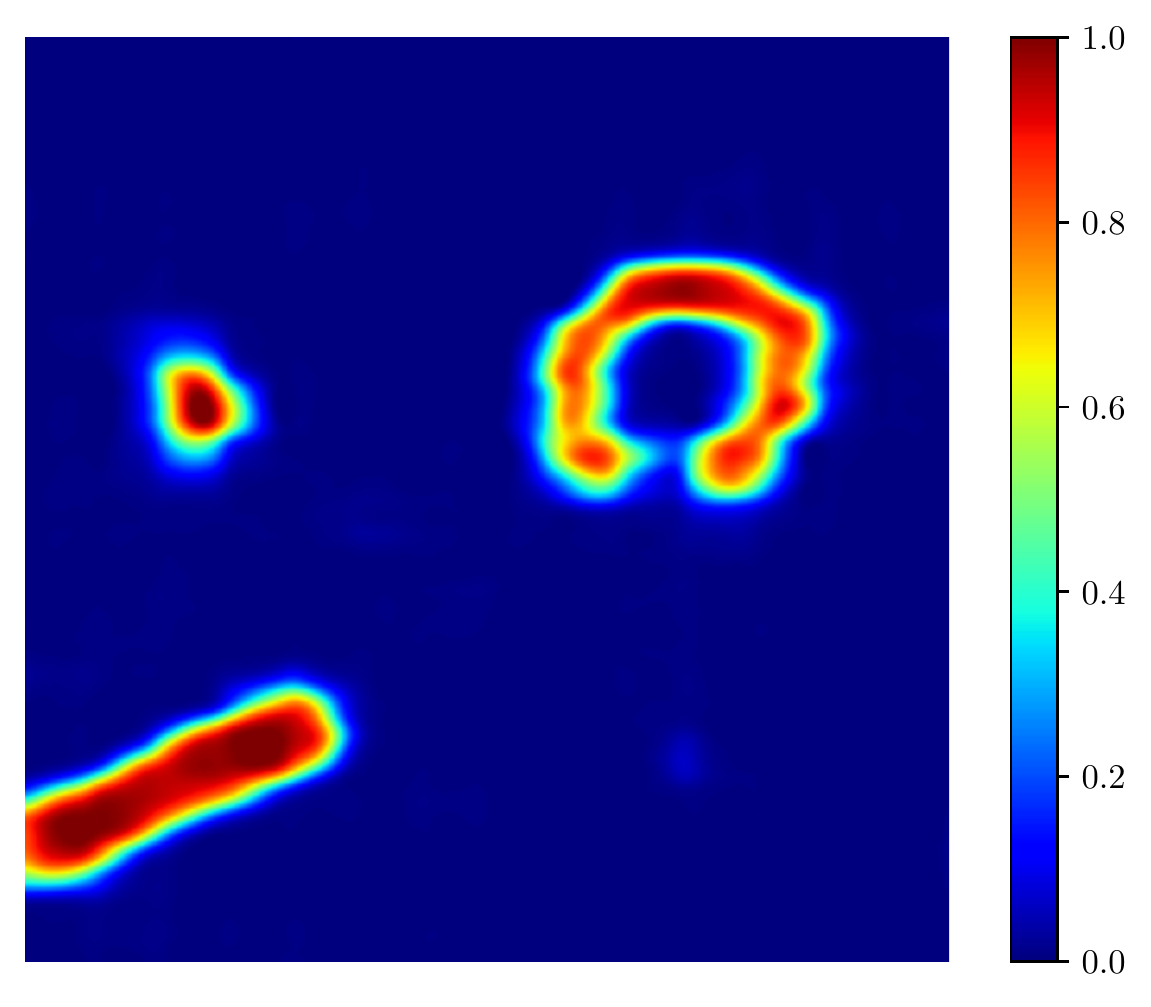}
		\includegraphics[width=0.077\textheight]{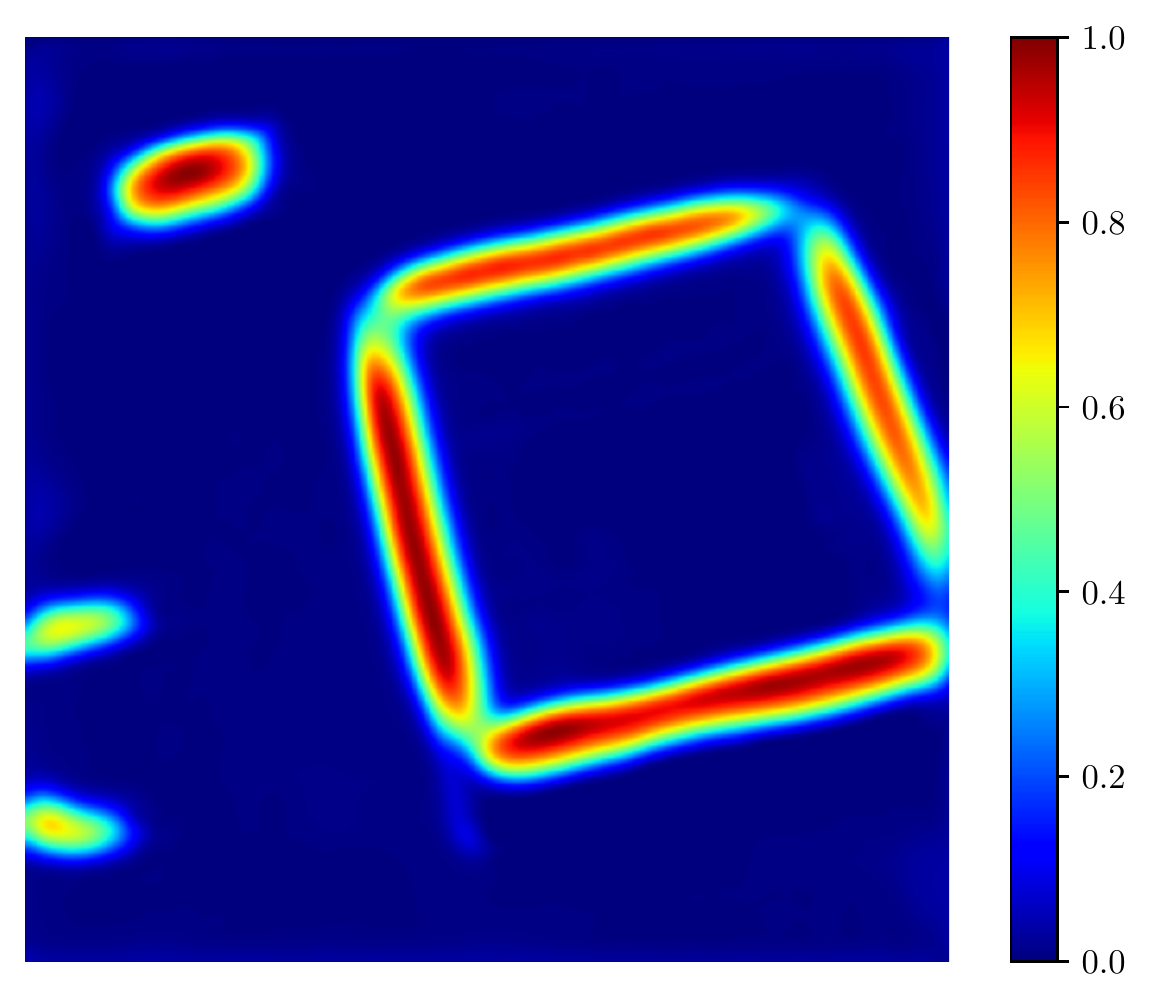}
		\includegraphics[width=0.077\textheight]{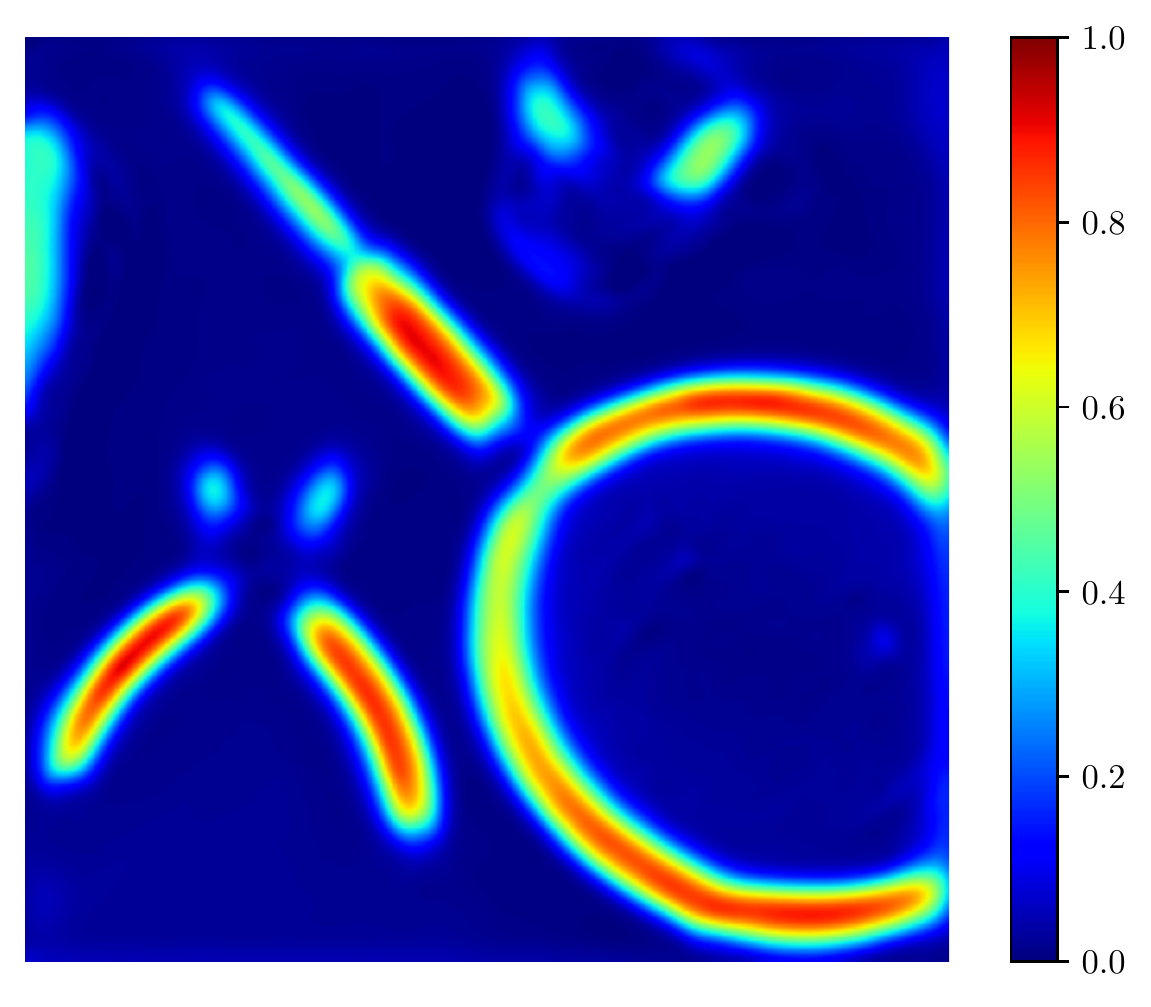}
		\includegraphics[width=0.077\textheight]{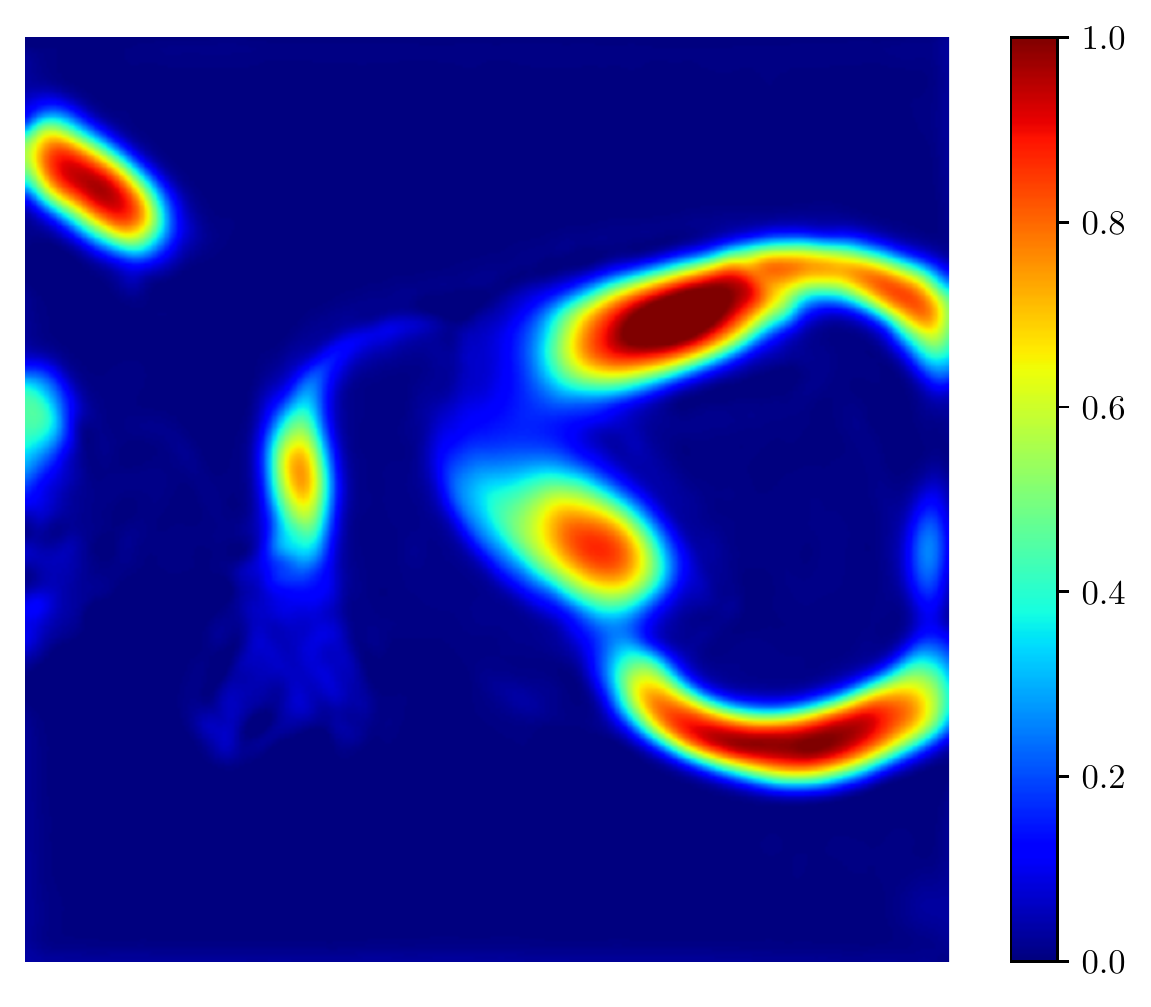}
		\includegraphics[width=0.077\textheight]{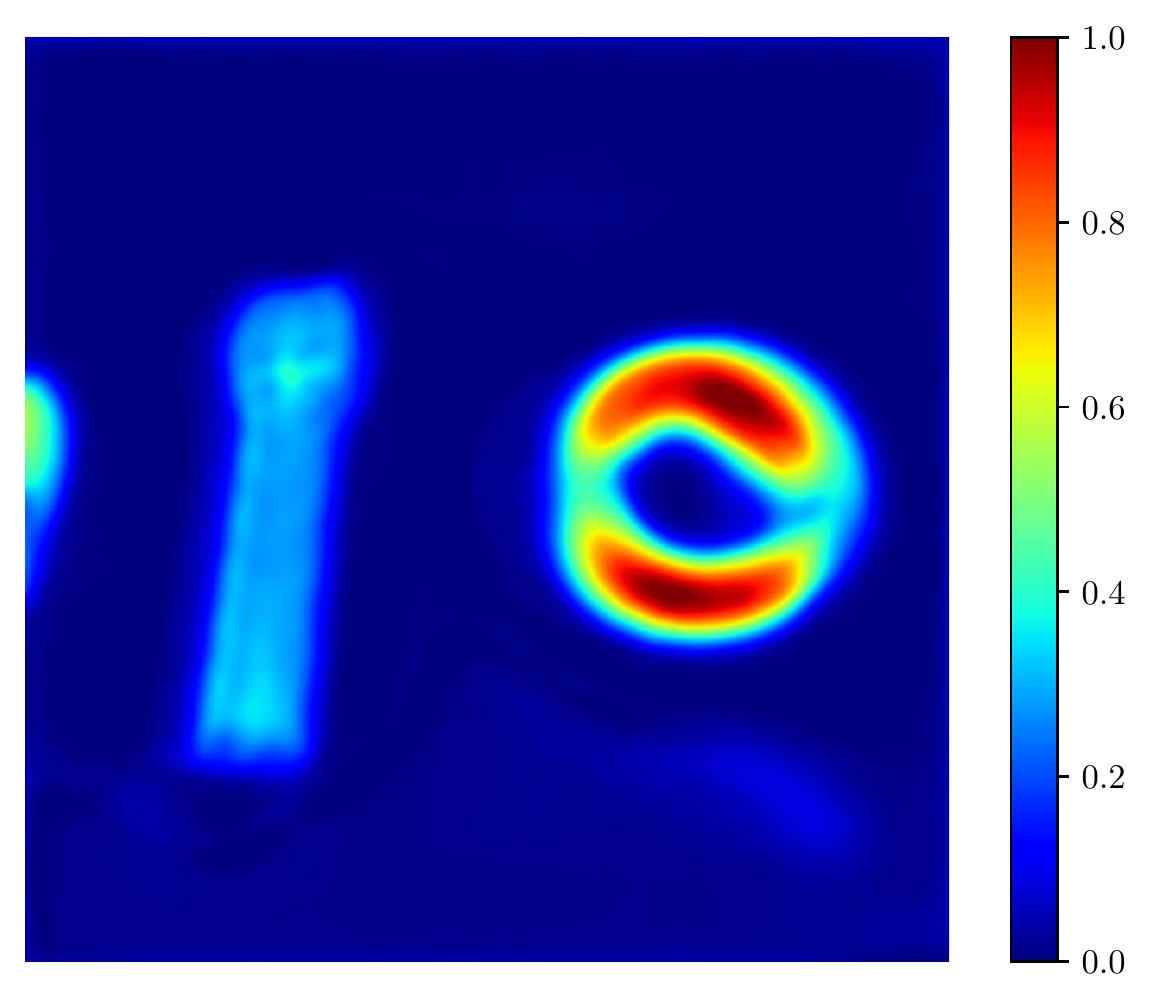}
		\includegraphics[width=0.077\textheight]{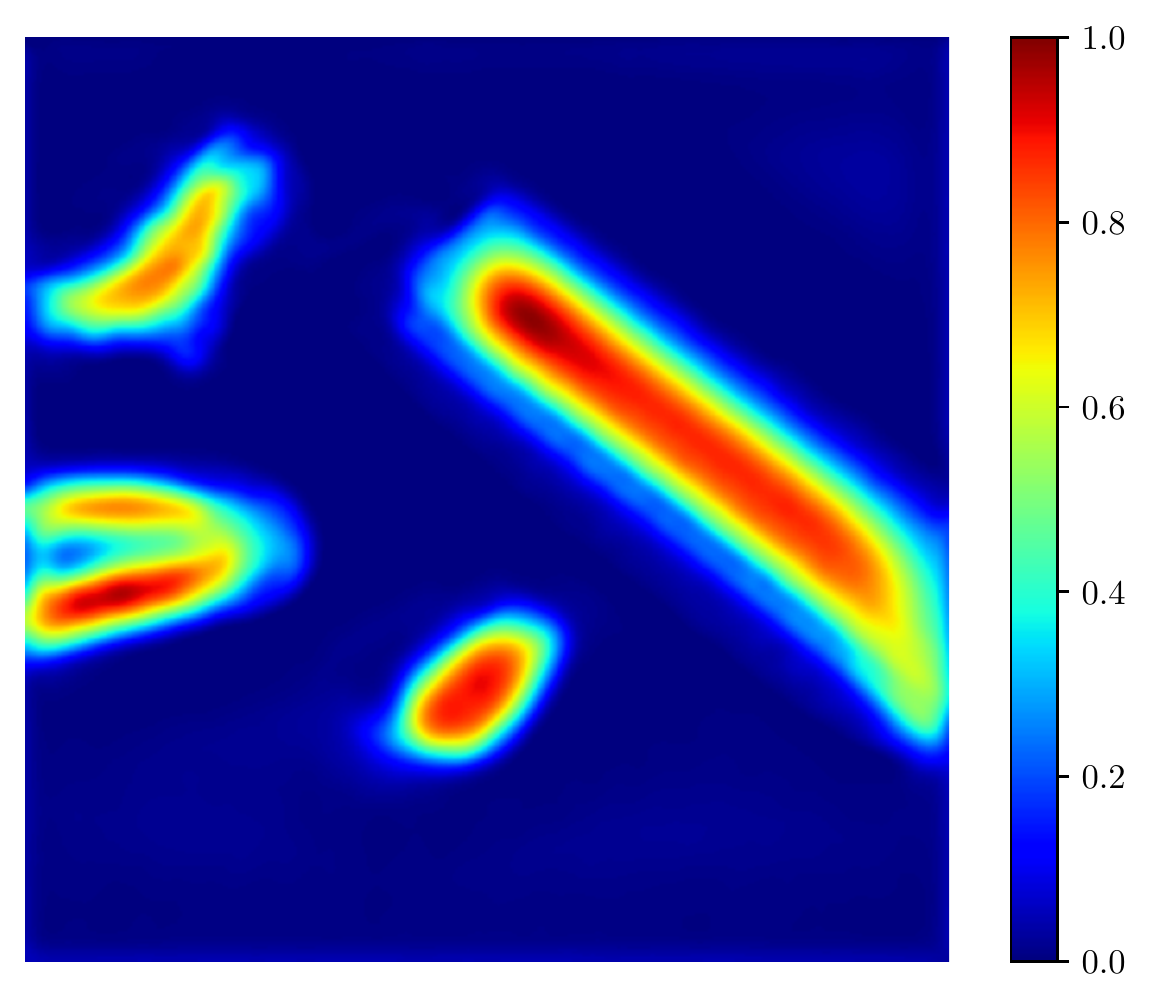}
	}
	\caption{Visualization comparison of the CNN and transformer-based grasping models.  }
	\label{attn_heatmap}	
\end{figure*}
{To better demonstrate whether the transformer-based grasping model can model the relationships between objects and across the scene, we present the multi-object  grasping results and grasping quality heatmaps of the transformer and CNN in Fig. \ref{attn_heatmap}. Our aim is to verify that the transformer is preferred over CNN for  visual grasping tasks and is better at capturing  global and local information. From Fig. \ref{attn_heatmap}, we can see that the grasp rectangles predicted by CNN have the right grasp position in most cases, but the predicted gripper angle and width are often not appropriate. In some cases, CNN even generates grasping rectangles in the  background. With the attention mechanism, our transformer-based model is able to clearly identify the objects from the background. 
	In the second row of Fig. \ref{attn_heatmap}, the grasping quality images show that the CNN-based approach can not identify the graspable area and consider the entire region of objects as a graspable zone with high success probabilities. Instead, as shown in the fourth row of Fig. \ref{attn_heatmap},  the transformer-based model is prone to capture the area that is easy to grasp due to its larger receptive field. For each attention block, the attention operation establishes the inter-element relationships through self-attention, and the subsequent multi-layer-perceptron (MLP) module further models the inherent relation between each element. The layer normalization and residual connections that interleave these two operations keep the training stable and efficient. In contrast, in CNN, the receptive field of each convolutional kernel is limited. To build a larger receptive field, the model often needs to repeatedly stack convolutional layers to gain global and semantically rich features. However, such a method in general results in the loss of detailed feature information such as the position and shape information of objects that are essential for grasping tasks. Therefore, we exploit a transformer-based model which can better capture not only the global information but also detailed features (e.g., the position and shape information). }
\subsection{Visualization Analysis}
\begin{figure*}[]
	
	\center
	\subfigure{
		\includegraphics[width=0.68\textheight]{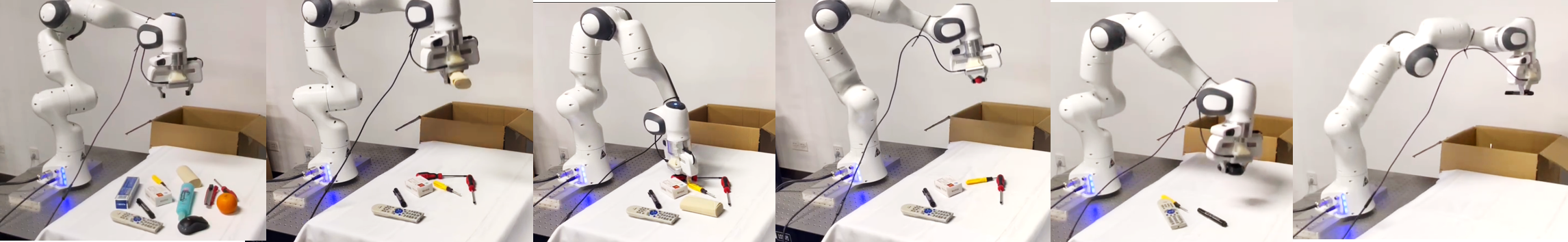}
	}
	\caption{ Screenshots of physical grasping in clutter. }	
	\label{grasp_show}
\end{figure*}

To clarify why the transformer architecture is helpful for grasp detection tasks, we visualize the heatmaps of attention maps, detailed in Fig. \ref{attention}.  From these heat maps, we can discover that the self-attention modules can readily learn the area that is easy for grasping,
such as the edges of objects, ignore irrelevant details, and pay more attention on the contour and shape of the objects. 
Meanwhile, the model focuses
on more general characteristics rather than individual
features. For example, for the chairs shown in Fig. \ref{attention},
our method evaluates the edge of the chairs with a higher
grasp quality. {We further provide more concrete examples of real-world grasping, and the experimental results show that the attention mechanism is more likely to achieve a better understanding of the grasping scenario, generate more accurate grasping rectangles, and work well on both household and novel objects.}
In Fig. \ref{grasp_show}, we illustrate a pick-and-place task based on our TF-Grasp on the Franka manipulator. Our grasp detection system works well for novel objects that have not been seen during training procedure and also locates graspable objects in cluttered environments.

In conclusion, the visualization results indicate that our TF-Grasp can produce a more  general and robust prediction, which contributes to  improving the detection accuracy.

\begin{table}[]
	\centering
	\setlength\tabcolsep{4pt}
	{
		\small
		\caption{Comparison between using and not using skip-connections }
		\begin{tabular}{c|c|c}
			\hline
			\multicolumn{3}{c}{The accuracy on Cornell Grasping Results} \\ \hline
			& With Skip-connections & Without Skip-connections  \\ \hline
			RGB & $96.78\%$& $95.7\%$\\ \hline
			Depth & $95.2\%$& $94.3\%$\\ \hline
			RGB+Depth & $97.99\%$& $96.1\%$\\ \hline
			\multicolumn{3}{c}{The accuracy on Jacquard Grasping Results} \\ \hline
			& With Skip-connections & Without Skip-connections  \\ \hline
			RGB & $93.57\%$& $92.4\%$\\ \hline
			Depth & $93.1\%$& $91.8\%$\\ \hline
			RGB+Depth & $94.6\%$& $93.27\%$\\ \hline

			%architecture &
			%  \multicolumn{4}{c|}{Encoder} &
			%  \\ \hline
		\end{tabular}
		\label{skip-connecetion}
	}
\end{table}
\subsection{Ablation Studies}
{To understand the role of skip-connections in our transformer model on the visual grasping problems, we conduct experiments on the Cornell and Jacquard grasping datasets with and without skip-connections using our transformer, respectively. The detailed experimental results are shown in Table \ref{skip-connecetion}.  The use of skip-connections is better than not using skip-connections in all input modes. The attention mechanism in the transformer builds inter-relationships in each layer, incorporates global features, and achieves  promising results. Through skip-connections, the multi-scale representations at different stages are further  fused globally. The empirical evidence shows that these further refinement and contextual features contribute to  the quality of final grasp prediction.  }

\subsection{Grasping in Real World Scenarios}

\textbf{Physical Setting.} The Franka Panda robot manipulation and the RealSense D435 RGB-D camera are used in our  physical experiment.  The camera is attached to the end-effector to keep  a good visual coverage of graspable objects.
In each grasp attempt, our TF-Grasp receives the visual signals from the depth camera mounted on the robot end-effector and outputs an optimal grasping posture. Next, the end-effector approaches the optimal target grasping posture based on the trajectory planned by a motion planning method, and then closes the gripper. Such a transformer-based grasp detection system can be easily adapted to other hardware platforms. During the grasp process, the raw depth sensor is filled with a portion of missing pixels that have NaN values. We  generate the mask of NaN values, normalize the depth image, and apply $cv2.inpaint$ \cite{bradski2000opencv} for further depth completion.

We perform a total of 165 grasping attempts, of which the robot performs successful grasp 152 times, achieving a success rate of $92.1\%$. Table \ref{real_grasp} lists the results of learning-based methods on real robot grasping. These results indicate that the transformer-based grasp detection system also behaves well on real robots.

\begin{table}[]
	\begin{center}
		\caption{The results for physical setup. } 
		\begin{tabular}{llc}
			
			\hline Authors & Physical  grasp & Success rate $(\%)$ \\
			\hline
			Lenz \cite{lenz2015deep} & 89/100 & 89\% \\
			Pinto \cite{pinto2016supersizing} & 109/150 & 73\% \\
			Morrison \cite{morrison2020learning} & 110/120 &92\% \\
			Chu\cite{chu2018real} &  89/100  & 89\%  \\

			\hline TF-Grasp(Ours)&  152/165  & $92.1\%$ \\

			\hline
		\end{tabular}
	\label{real_grasp}
	\end{center}
\end{table}

\section{Discussion and Conclusion}
In this work, we develop a novel architecture for visual grasping. 
Although CNN and its variants are still the dominant models in visual robotic grasping, we show the powerful potential of transformers in grasp detection. Compared with CNN-based  counterparts, the transformer-based grasp detection models are better at capturing global dependencies and learning powerful feature representation.
The results show that our proposed approach outperforms original CNN-based models. The contexts can be better represented by attention propagation. Nevertheless, the current approach is limited to the parallel gripper. Future research will focus on developing a universal transformer-based grasp detection method for other types of grippers, such as the five finger dexterous hand.


\ifCLASSOPTIONcaptionsoff
  \newpage
\fi

\end{document}